\RequirePackage{fix-cm}

\documentclass[twocolumn]{svjour3}          

\smartqed  
\usepackage{epsfig}
\usepackage{graphicx}
\usepackage{amsmath}
\usepackage{amssymb}
\usepackage{subfigure}
\usepackage{algorithmic}
\usepackage{algorithm}
\usepackage{url}
\usepackage{multirow}
\usepackage{booktabs}
\usepackage{xcolor}
\usepackage{enumerate}
\usepackage[most]{tcolorbox}

\usepackage{ragged2e} 
\usepackage{xspace}
\usepackage{bm}
\makeatletter
\def\UrlAlphabet{%
      \do\a\do\b\do\c\do\d\do\e\do\f\do\g\do\h\do\i\do\j%
      \do\k\do\l\do\m\do\n\do\o\do\p\do\q\do\r\do\s\do\t%
      \do\u\do\v\do\w\do\x\do\y\do\z\do\A\do\B\do\C\do\D%
      \do\E\do\F\do\G\do\H\do\I\do\J\do\K\do\L\do\M\do\N%
      \do\O\do\P\do\Q\do\R\do\S\do\T\do\U\do\V\do\W\do\X%
      \do\Y\do\Z}
\def\UrlDigits{\do\1\do\2\do\3\do\4\do\5\do\6\do\7\do\8\do\9\do\0}
\g@addto@macro{\UrlBreaks}{\UrlOrds}
\g@addto@macro{\UrlBreaks}{\UrlAlphabet}
\g@addto@macro{\UrlBreaks}{\UrlDigits}

\usepackage[pagebackref=true,breaklinks=true,letterpaper=true,colorlinks,bookmarks=false]{hyperref}

\newcommand{\tabincell}[2]{\begin{tabular}{@{}#1@{}}#2\end{tabular}} 

\newcommand{\etal}{{\it et al.\,}}
\usepackage{xcolor}
\usepackage{color}
\definecolor{gray}{rgb}{0.5,0.5,0.5}

\newcommand{\added}[1]{\textcolor{black}{#1} }
\newcommand{\revision}[1]{\textcolor{black}{#1} }

\usepackage{amssymb}
\usepackage{pifont}
\newcommand{\cmark}{\ding{51}}%
\newcommand{\xmark}{\ding{55}}%

\makeatletter
\DeclareRobustCommand\onedot{\futurelet\@let@token\@onedot}
\def\@onedot{\ifx\@let@token.\else.\null\fi\xspace}
\def\eg{{\em e.g}\onedot} 
\def\ie{{\em i.e}\onedot}

\def\vs{{\em vs}\onedot}
\def\wrt{{\em w.r.t\onedot}}

\def\etal{{\em et al}\onedot}
\def \x {\mathbf{x}}
\def \M {\mathcal M}

\def\modelname{Semantic Representational Distillation}
\def\shortname{SRD}

\begin{document}

\title{Knowledge Distillation Meets Open-Set Semi-Supervised Learning
}

\author{Jing Yang         \and
        Xiatian Zhu    \and
        Adrian Bulat  \and
        Brais Martinez    \and
        Georgios Tzimiropoulos
}

\institute{
           Jing Yang \at
           University of Cambridge, UK. \\
           \email{y.jing2016@gmail.com}           \\
           \and
           Xiatian Zhu \at
           University of Surrey, Guildford, UK. \\
           \and
           Adrian Bulat  \at
           Samsung AI Centre, Cambridge, UK.\\
           \and
           Brais Martinez  \at
           Samsung AI Centre, Cambridge, UK.\\
           \and
           Georgios Tzimiropoulos \at
           Samsung AI Centre, Cambridge, UK.\\
           Queen Mary University London, London, UK\\
}

\date{Received / Accepted date}

\maketitle

\begin{abstract}
\justifying 
Existing knowledge distillation methods mostly focus on distillation
of teacher's prediction and intermediate activation. However, the structured representation, which arguably is one of the most critical ingredients of deep models, is largely overlooked. In this work, we propose a novel {\em \modelname{}} ({\bf\em \shortname{})} method dedicated for distilling representational knowledge semantically from a pretrained teacher to a target student. The key idea is that we leverage the teacher's classifier as a semantic critic for evaluating the representations of both teacher and student and distilling the semantic knowledge with high-order structured information over all feature dimensions. This is accomplished by introducing a notion of cross-network logit computed through passing student's representation into teacher's classifier. Further, considering the set of seen classes as a basis for the semantic space in a combinatorial perspective, we scale \shortname{} to unseen classes for enabling effective exploitation of largely available, arbitrary unlabeled training data. At the problem level, this establishes an interesting connection between knowledge distillation with open-set semi-supervised learning (SSL). Extensive experiments show that our \shortname{} outperforms significantly previous state-of-the-art knowledge distillation methods on both coarse object classification and fine face recognition tasks, as well as less studied yet practically crucial binary network distillation. Under more realistic open-set SSL settings we introduce, we reveal that knowledge distillation is generally more effective than existing Out-Of-Distribution (OOD) sample detection,
and our proposed \shortname{} is superior over
both previous distillation and SSL competitors.
The source code is available at  \url{https://github.com/jingyang2017/SRD\_ossl}.


\end{abstract}

\section{Introduction}
\label{section:Introduction}
Optimizing lightweight Convolutional Neural Networks (CNNs) to be highly performing is critical,
\eg, enabling the developments on resource-limited platforms such as mobile devices.
To that end, different model compression approaches have been extensively investigated, including network pruning~\cite{han2015deep,lebedev2016fast}, 
network quantization~\cite{rastegari2016xnor,wu2016quantized}, 
neural architecture search~\cite{zoph2016neural,liu2018darts},
and knowledge distillation~\cite{hinton2015distilling,zagoruyko2016paying}.
In particular, knowledge distillation aims to transfer the knowledge from a stronger network (\ie, the teacher) to another (\ie, the student). 
Typically, the teacher is a high-capacity model or an ensemble capable of achieving stronger performance, while the student is a compact model with much fewer parameters and requiring much less computation.
The {\em objective} is to facilitate the optimization of the student by leveraging the teacher's capacity. 
A general rationale behind distillation can be explained from an optimization perspective that higher-capacity models are able to seek for better local minima thanks to over-parameterization~\cite{du2018power,soltanolkotabi2018theoretical}. 

Existing knowledge distillation methods 
start with transferring classification predictions~\cite{hinton2015distilling}
and intermediate representations (\eg, feature tensors~\cite{romero2015fitnets} and attention maps~\cite{zagoruyko2016paying}).
However, they suffer from a limitation of distilling {\em structured} representational knowledge including the latent complex interdependencies and correlations between different dimensions. 
This is because their objective formulations typically treat all the feature or prediction dimensions {\em independently}.
Motivated by this analysis, a representation distillation method \cite{tian2019contrastive} is recently developed by contrastive learning~\cite{chen2020simple,he2020momentum}.  
The concrete idea is to maximize the representation's mutual information 
across the teacher and student via contrastive learning.
Despite a principled solution following seminal information theory \cite{infotheorybook2006thomas}, this method is limited in high-level semantics perception and distillation. 
Because the teacher's classifier,
that maps the feature representation to the semantic class space, is totally ignored during distillation.
Further, contrastive learning often requires a large number of training samples in loss computation, meaning a need of resource demanding large mini-batch or complex remedy (\eg, using a memory bank).

To overcome the aforementioned limitations, in this work a novel {\bf\em \modelname{}} (\shortname{}) is introduced.
Our key idea is to leverage the pretrained teacher's classifier as a {\em semantic critic} for guiding representational distillation in a classification-aware manner.
Concretely, we introduce a notion of cross-network logit,
obtained by feeding the student's representation to the teacher's classifier. 
Subject to the teacher and student sharing the same input, aligning the cross-network logit with the teacher's counterpart can then enable the distillation of high-order semantic correlations among feature dimensions, \ie, semantic distillation of representation. 
Further, we extend the proposed \shortname{} to open-set semi-supervised learning (SSL) by exploiting unconstrained unlabeled data from arbitrary classes.
This is motivated by our perspective that seen classes of labeled training data can be regarded {\em collectively} as a basis of the semantic space in the linear algebra theory; And any unseen classes can be approximated by a specific combination of seen classes.
This hypothesis naturally breaks the obstacles of generalizing the knowledge of seen classes to unseen classes, a key underlying challenge in solving open-set SSL (\eg, over-confident classification on the samples of unseen classes \cite{chen2020semi}).

Our {\bf contributions} are three-fold:
{\bf (I)} We propose a simple yet effective {\em \modelname} (\shortname{}) method with a focus on structured representation optimization via semantic knowledge distillation.
This is realized by taking the teacher's classifier as a semantic critic used for evaluating both teacher and student's representation in terms of their classification performance and ability.
{\bf (II)} We connect {semantic distillation} with open-set semi-supervised learning based on an idea that seen classes can be used as a basis of the semantic space.
{\bf (III)} Extensive experiments show that the proposed \shortname{} method can train more generalizable student models than the state-of-the-art distillation methods across a variety of network architectures (\eg, Wide ResNets, ResNets, and MobileNets) and recognition tasks (\eg, coarse-grained object classification and fine-grained face recognition, real and binary network distillation).
Compared to previous open-set SSL works, we further introduce more realistic experiment settings characterized 
by more classes and unlabeled data with different distributions, as well as less common classes between labeled and unlabeled sets.
Critically, our experiments reveal that knowledge distillation turns out to be a more effective strategy than previously often adopted Out-Of-Distribution (OOD) detection (Table \ref{tab:un_kds} \vs{} Table \ref{tab:ssl_tin}),
and our \shortname{} outperforms both state-of-the-art distillation and SSL methods, often by a large margin.
On the other hand, it is also shown that OOD detection brings very marginal benefits to knowledge distillation methods (Table \ref{tab:KD_OOD}).

This is an extension of our preliminary ICLR 2021 work \cite{srrl}.
We further make the following significant contributions:
{\bf (1)} Extending our method in general knowledge distillation to open-set semi-supervised learning, two previously independently investigated fields.
{\bf (2)} Analyzing the limitations of existing open-set SSL settings and introducing more realistic ones with less constrained unlabeled data such as less class overlap between labeled and unlabeled sets. 
{\bf (3)} Evaluating and comparing comprehensively both knowledge distillation and open-set SSL methods with new findings and insights in tackling more unconstrained unlabeled data.
{\bf (4)} To show the generality of our approach, we evaluate on a diverse range of problems with varying underlying characteristics, such as coarse-grained object classification tasks and fine-grained face recognition distillation.

\section{Related Work}
\label{section:Related Work}

\subsection{Knowledge Distillation} 
Knowledge distillation is an effective approach to 
optimizing low-capacity networks with extensive studies 
in image classification \cite{hinton2015distilling,romero2015fitnets,zagoruyko2016paying,huang2017like,yim2017gift,zhang2018deep,zhou2018rocket,lee2018self,tung2019similarity,kim2018paraphrasing,lan2018knowledge,peng2019correlation,Cho_2019_ICCV,park2019relational,liu2019knowledge,heo2019knowledge,heo2019comprehensive}.
Existing distillation methods can be generally divided into two categories: 
isolated knowledge based and relational knowledge based.

\vspace{0.1cm}
\noindent
\textbf{Isolated knowledge based methods: } 
The seminal work in Hinton et al. \cite{hinton2015distilling} popularized the research on knowledge distillation by simply distilling the teacher's classification outputs (\ie, the knowledge).
Compared to one-hot class label representation, this knowledge is semantically richer due to involving underlying inter-class similarity information.
Soon after, intermediate teacher representations such as feature tensors~\cite{romero2015fitnets}
are also leveraged for richer distillation.
However, matching the whole feature tensors is not necessarily viable in certain circumstances due to the capacity gap between the teacher and student, even adversely affecting the performance and convergence of the student.
As an efficient remedy, Attention Transfer (AT)~\cite{zagoruyko2016paying} might be more achievable as the feature attention maps (\ie, a summary of  all the feature channels) represent a more flexible knowledge to be learned.
The following extended AT based on the maximum mean discrepancy of the activations~\cite{huang2017like} shares the same spirit. 
Interestingly, Cho \etal{} \cite{Cho_2019_ICCV} reveal that very highly strong networks would be ``too good'' to be effective teachers. To mitigate this issue, they early stop the teacher's training. 
Later on, Heo \etal{}~\cite{heo2019comprehensive} study the effect of distillation location within the network, along with margin ReLU and a specifically designed distance function for maximizing positive knowledge transfer.
More recently, Passalis \etal{} \cite{passalis2020heterogeneous} leverage the previously ignored information plasticity by exploiting information flow through teacher's layers. 

\vspace{0.1cm}
\noindent
\textbf{Relational knowledge based methods:}
Another line of knowledge distillation methods instead explore relational knowledge.
For example, Yim \etal{} \cite{yim2017gift} distil the feature correlation
by aligning the layer-wise Gram matrix of feature representations across the teacher and student. A clear limitation of this method is at a high computational cost. This could be alleviated to some extent by compressing the feature maps using singular value decomposition~\cite{lee2018self}. Park \etal{} \cite{park2019relational} consider both distance-wise and angle-wise relations of each embedded feature vector. 
This idea is subsequently extended by~\cite{peng2019correlation}
for better capturing the correlation between multiple instances with Taylor series expansion, and by \cite{liu2019knowledge} for modeling the feature space transformation across layers via a graph with the instance feature and relationship as vertexes and edges.
Inspired by an observation that semantically similar samples should give similar activation patterns, Tung \etal{} \cite{tung2019similarity} introduce an idea of similarity-preserving knowledge distillation \wrt{} the generation of either similar or dissimilar activations.
Besides, Jain \etal{} \cite{jain2019quest} exploit the relational knowledge \wrt{} a quantized visual word space during distillation. 
For capturing more detailed and fine-grained information,
Li \etal{} \cite{lilocal} employ the relationship among local regions in the feature space.
In order to distil richer representational knowledge from the teacher,
Tian \etal{} \cite{tian2019contrastive} maximize the representation's mutual information between the teacher and student.
%
Whilst sharing a similar objective, in this work we instead correlate the teacher's and student's representations via considering the pretrained teacher's classifier as a semantic critic.

Despite its simplicity, we show that our method is superior and more generalizable than prior work \cite{tian2019contrastive} in distilling the underlying semantic representation information over a variety of applications (see Sec. \ref{sec:sup_sota} and Sec. \ref{sec:unsup_kd}). 
%

Different from previous works, we further exploit the potential of distillation using unlabeled training data often available at scale. 
This brings together the two fields of knowledge distillation and open-set semi-supervised learning \cite{oliver2018realistic,chen2020semi}, both of which develop independently, 
and importantly presents a unified perspective and common ground that enable natural model comparison and idea exchange across the two fields. 

\subsection{Open-Set Semi-Supervised Learning} 
\label{sec:ssl}

Most existing semi-supervised learning (SSL) works \cite{sohn2020fixmatch,lee2013pseudo,tarvainen2017mean,berthelot2019mixmatch,shi2018transductive,iscen2019label,sajjadi2016regularization,laine2016temporal,miyato2018virtual}
make a {\em closed-set} assumption that unlabeled training data share the same label space as the labeled data. 
This assumption, however, is highly artificial and may hinder the effectiveness of SSL when processing real-world unconstrained unlabeled data with unseen classes, \ie, out-of-distribution (OOD) samples \cite{oliver2018realistic}. 
This is because OOD data could cause harmful error propagation,
\eg, via incorrect pseudo labels.

\revision{To further generalize SSL to unconstrained data without labels, there is a recent trend of developing more realistic \textit{open-set} SSL methods \cite{chen2020semi,guo2020safe,yu2020multi,huang2021universal,T2T,rizve2022towards,openmatch}}.
A common strategy of these works is to identify and suppress/discard OOD samples as they are considered to be less/not beneficial. 
Specifically, pioneer methods (UASD \cite{chen2020semi} and DS$^3$L \cite{guo2020safe}) leverage a dynamic weighting function based on the OOD likelihood of an unlabeled sample.
Curriculum learning has been used to detect and drop potentially detrimental data \cite{yu2020multi}. 
Besides, T2T \cite{T2T} pretrains the feature model with all unlabeled data for improving OOD detection. 
\added{\cite{rizve2022towards} leverages both sample uncertainty and prior knowledge about class distribution to produce pseudo-labels for unlabeled data.}
More recently, OpenMatch \cite{openmatch} trains a set of one-vs-all classifiers for OOD detection and removal during SSL.

Whilst taking a step away from the artificial closed-set assumption, most existing open-set SSL works either focus on a simplified setting where both labeled and unlabeled sets are drawn from a single dataset, or consider only limited known classes and unlabeled data \cite{T2T,yu2020multi} with all the known classes included in the unlabeled set \cite{openmatch,T2T}.
Clearly, both cases are fairly ideal and hardly valid in many practical cases.
To overcome this limitation, we introduce more realistic open-set SSL settings characterized by more classes and unlabeled data with distinct distributions, and less class overlap between labeled and unlabeled sets.
Critically, we find that existing open-set SSL methods fail to benefit from using unlabeled data under such unconstrained settings (see Table \ref{tab:ssl_tin}).
This challenges all the previous OOD based findings and is thought-provoking.
The main reasons we find include more challenging OOD detection and the intrinsic limitation of exploiting unlabeled samples from seen classes alone.
Further, our experiments show that knowledge distillation methods provide a more effective and reliable solution to leverage unlabeled data with less constraints (see Table \ref{tab:un_kds} \vs{} Tables \ref{tab:ssl_tin}, \revision{\ref{tab:KD_place}, and \ref{tab:KD_cc3m}}).

Self-supervised learning has been leveraged for enhancing 
knowledge distillation \cite{SSKD}.
Interestingly, the usage of unlabeled data is not considered.
We empirically show that our \shortname{} can readily benefit from this strategy in the open-set SSL setting (Table \ref{tab:sskd}).


\begin{figure*}[ht]
\centering
\includegraphics[width=0.98\textwidth,trim={0.2cm 0.cm 0cm 0.cm},clip]{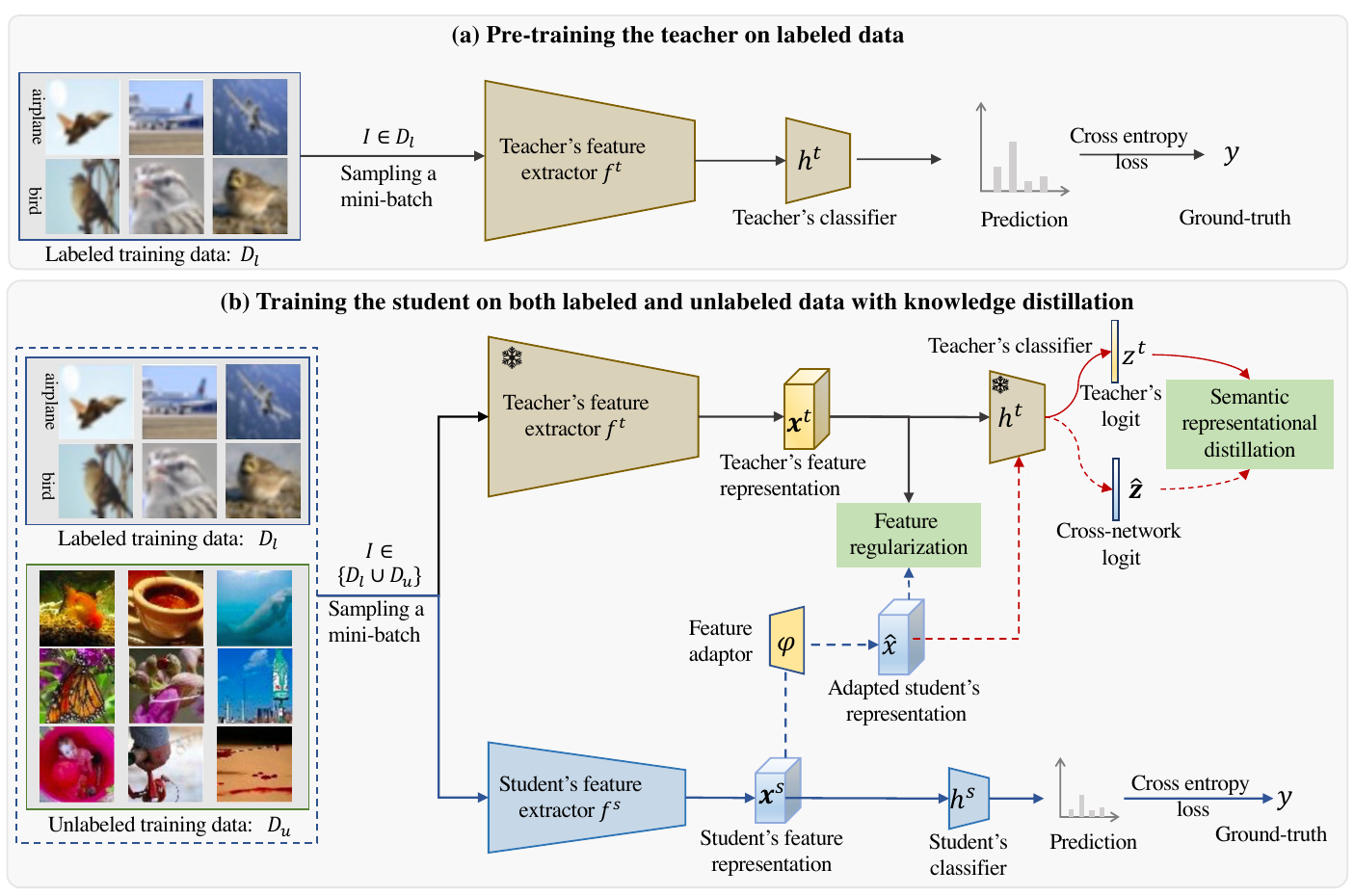}
\caption{Schematic overview of the proposed {\em \modelname} (\shortname) method for knowledge distillation at the presence of both labeled and unlabeled data.
Following the knowledge distillation pipeline, {\bf(a)} we first pretrain a teacher model on the labeled training set.
{\bf(b)} Subsequently, we distil the semantic knowledge from the pretrained frozen teacher to improve the optimization of a student.
Specifically, given a training image $I$, we feed it into both the teacher $T$ and the student $S$ to obtain the feature representations $\bm{x}^t$ and $\bm{x}^s$. 
Critically, we introduce a notion of {\em cross-network} logit $\hat{\bm{z}}$, obtained by passing the student's representation $\bm{x}^s$
into the teacher's classifier $h^t$ via a feature adaptor $\varphi$.
Considering the teacher's classifier $h^t$ as a semantic critic, 
we distil the semantic knowledge of $\bm{x}^t$ to $\bm{x}^s$
via aligning the {cross-network} logit $\hat{\bm{z}}$ towards to
the teacher's logit $\bm{z}^t$.
In this design, the two representations $\bm{x}^t$ and $\bm{x}^s$
share the same semantic critic (\ie, classifier) which could
facilitate representational knowledge distillation.
To further ease the semantic distillation, we impose a feature-level alignment regularization $\mathcal{R}$ between the teacher's representation $\bm{x}^t$ and the adapted student's representation $\hat{\bm{x}}$.
For labeled training samples, we also apply a supervised learning supervision on the student's prediction.
}
\label{fig:overview}
\end{figure*}

\section{Method}
A generic CNN consists of a feature extractor $f: I \rightarrow \bm{x}$,
and a classifier $h: \bm{x} \longrightarrow \bm{p}$,
where $I \in \mathcal{R}^{H\times W \times 3}$, $\bm{x} \in \mathcal{R}^d$, $\bm{p} = [p_1, \cdots, p_k, \cdots, p_K] \in \mathcal{R}^K$ denote an input image sized at $H \times W$,
its feature vector of $d$ dimensions, and its classification probability over $K$ classes, respectively.
Often, $\bm{x}$ is obtained by global average pooling over the last feature map $\bm{F}$.
The classifier $h$ is parameterized by a projection matrix $\bm{W} \in \mathbb{R}^{d \times K}$ that first projects $\bm{x}$ into the logits:
$\bm{z} = \bm{W}^{\top} \bm{x} = [z_1, \cdots, z_k, \cdots, z_K] $, followed by {softmax} normalization: 
\begin{align}\label{eq:sm}
    {p}_k = \texttt{sm}({z}_k)=\frac{\exp({z}_k)}{\sum_{k'=1}^K{\exp({z}_{k'})}},
\end{align}
where $k \in \{1,\cdots,K\}$ indexes the class. 

\vspace{0.1cm}
In general {\em knowledge distillation}~\cite{hinton2015distilling}, we have a teacher network $T=\{f^t, h^t\}$ and a student (target) network $S=\{f^s, h^s\}$. 
It has two steps in training. 
In the first step, the teacher network $T$ is pretrained on a labeled training set $\mathcal{D}_l$ in a supervised learning manner.
Often, the cross entropy loss is adopted as
\begin{align}\label{eq:ce}
\mathcal{L}_{ce} = -\sum_{k=1}^{K} y_k \log p_k
\end{align}
where $\bm{y} = [y_1, \cdots, y_k, \cdots, y_K] \in \mathcal{Y}$ 
is the one-hot ground-truth label of a given input image $I \in \mathcal{D}_l$.
In the second step, the student network is then trained under distillation with the frozen teacher and supervised learning with the ground-truth labels (\eg, the cross-entropy loss).
A typical knowledge distillation process is realized by logit-matching \cite{hinton2015distilling} that minimizes the KL divergence between the logits of $T$ and $S$ as:
\begin{align}\label{eq:kd}
\mathcal{L}_{kd} = -\sum_{k=1}^{K} p_k^t \log p_k^s, \;\; \text{where} \\ \nonumber
\bm{p}^t = [p_1^t, \cdots, p_K^t] = \texttt{sm}(\bm{z}^{t}), \bm{z}^{t} = h^t(\bm{x}^t), \bm{x}^t = f^t(I);
\\ \nonumber
\bm{p}^s = [p_1^s, \cdots, p_K^s]  = {\texttt{sm}(\bm{z}^{s})}, \bm{z}^{s} = h^s(\bm{x}^s),\bm{x}^s = f^s(I).
\end{align}
Whilst this formula has shown to be effective,
we consider it is less dedicated on distilling teacher's representation knowledge, especially when considering the structured correlations and inter-dependencies between distinctive feature dimensions.

\subsection{\modelname}
\label{sec:method_SRD}
To overcome the aforementioned problem, we propose a novel distillation method, dubbed as
{\bf \em \modelname} (\shortname), dedicated to enhancing the representational transfer from the pretrained teacher to the target student
in the distillation process.
An overview of \shortname{} is depicted in Fig.~\ref{fig:overview}.
Specifically, we leverage the pretrained teacher's classifier $h^t$ as a {\em semantic critic} for explicitly distilling the underlying semantic knowledge of the teacher's representation $\bm{x}^t$ to the student's counterpart $\bm{x}^s$.
That is, the same classifier is shared by the two representations $\bm{x}^t$ and $\bm{x}^s$ for facilitating representational knowledge distillation via a dedicated channel.

Formally, we pass the student's representation $\bm{x}^{s}$ through the teacher's classifier $h^{t}$ to obtain the {\bf \em cross-network logit} as:
\begin{align} \label{eq:xnet_logit}
    \hat{\bm{z}} = h^t \big( \varphi(\bm{x}^s) \big) = [\hat{z}_1, \cdots, \hat{z}_k, \cdots, \hat{z}_K],
\end{align}
where $\varphi$ is a representation adaptor
for making $\bm{x}^s$ compatible with the teacher's classifier.
In practice, $\varphi$ is implemented by a 1$\times$1 convolutional layer with batch normalization and activation applied on the last feature map of the student.
We formulate a general \shortname{} objective function as: 
\begin{align}\label{eq:srd}
    \mathcal{L}_{srd} = \texttt{dist}(\bm{z}^t, \hat{\bm{z}}),
\end{align}
where the teacher's logit $\bm{z}^t$, and $\texttt{dist}(\cdot, \cdot)$ denotes
any distance metric.
For an extreme case of $\bm{z}^t = \hat{\bm{z}}$ (corresponding to the minimal $\mathcal{L}_{srd}$), since the teacher's classifier is shared by both representations, this means that $\bm{x}^s=\bm{x}^t$ subject to some sufficient conditions such as full rank transformation matrix, \ie, the full knowledge of $\bm{x}^t$ has been transferred to $\bm{x}^s$.
Generally, minimizing $\mathcal{L}_{srd}$ is equivalent to maximizing the knowledge transfer from $\bm{x}^t$ to $\bm{x}^s$.

\vspace{0.1cm}
\noindent{\bf \shortname{} objective instantiation:}
To implement \shortname{} objective,
we consider three different designs.
The {\bf \em first} design adopts the KL divergence, following the logit-matching distillation function as:
\begin{equation}\label{eq:srd_kl}
\mathcal{L}_{srd}^{kl} =  -\sum_{k=1}^{K} p_k^t \log \hat{p}_k,
\end{equation}
where $\bm{\hat{p}} = [\hat{p}_1, \cdots, \hat{p}_k, \cdots, \hat{p}_K] = \texttt{sm}(\hat{\bm{z}})$ is the cross-network classification probability
and $p_k^t$ is the teacher's classification probability  obtained as in Eq. \eqref{eq:sm}.
It is noteworthy that, the logit-matching distillation (Eq. \eqref{eq:kd}) uses specific classifiers $h^t$/$h^s$ for the representations $\bm{x}^t$/$\bm{x}^s$, separately;
Compared to our \shortname{} sharing a single classifier for both representations, this gives more degrees of freedom to the optimization of feature extractor,
resulting in less dedicated constraint on representational knowledge distillation.
We will show in the experiments (Sec.~\ref{sec:ablation}) that
our \shortname{} can yield clearly superior generalization capability.

In the {\bf \em second} design, we adopt the mean square error (MSE) as the distillation loss:
\begin{align}\label{eq:srd_mse}
\mathcal{L}_{srd}^{mse}  = 
\left \| \bm{z}_t -\hat{\bm{z}} \right\|^2
=
 \left \| (\bm{W}^t)^\top (\bm{x}^t-\varphi(\bm{x}^s)) \right\|^2.
\end{align}
This is essentially a Mahalanobis distance with the linear transformation defined by the teacher's classifier weights $\bm{W}^t$.
As is pretrained, this imposes semantic correlation over all the feature dimensions, making the distillation process class discriminative. 

In the {\bf \em third} design, we consider classification probability MSE by further applying the softmax normalization as:
\begin{equation}\label{eq:srd_msev2}
\mathcal{L}_{srd}^{pmse} = \left \| \bm{p}^t -\bm{\hat{p}} \right\|^2.
\end{equation}
This allows us to evaluate the effect of normalization in comparison to the {\bf \em second} design.
We will evaluate these different designs in our experiment (Table \ref{tab:ab_srd}).

\vspace{0.1cm}
\noindent{\bf Overall objective:}
We formulate the overall loss objective function of \shortname{} on the labeled training set $\mathcal{D}_l$
as:
\begin{equation}
\mathcal{L}_l = \mathcal{L}_{ce}(\mathcal{D}_l) + 
\alpha \mathcal{L}_{srd}(\mathcal{D}_l) + 
\beta \mathcal{R}(\mathcal{D}_l), \label{eq:allloss}
\end{equation}
where $\mathcal{L}_{ce}$ is the cross-entropy loss computed between the student's classification probability $\bm{p}^s$ and ground-truth labels, as defined in Eq. \eqref{eq:ce}.
$\mathcal{R} = \| \bm{x}^t - \varphi(\bm{x}^s)\|$ is a feature regularization inspired by the notion of feature matching of FitNets~\cite{hinton2015distilling}.
This is conceptually complementary with \shortname{} loss $\mathcal{L}_{srd}$ 
as it functions directly in the representation space and
potentially facilitates the convergence of our \shortname{} loss.
The two scaling parameters $\alpha$ and $\beta$ control the impact of respective loss terms. 

\vspace{0.1cm}
\subsection{Meeting Open-Set Semi-Supervised Learning}



The aim of knowledge distillation is to transfer the learned knowledge from the teacher to the student.
The standard setting is to exclusively use the labeled training set of a \textit{target domain} to train the student. 
It is a supervised learning scenario.
However, distillation methods \cite{hinton2015distilling,romero2015fitnets} typically require no ground-truth labels,
presenting an unsupervised learning property.
Therefore, only using the labeled training set of 
target domain is unnecessarily restricted
and extra unlabeled data should be well incorporated for improved knowledge distillation.
Technically, inspired by the spirit of linear algebra,
we consider the appearential characteristics of unseen classes can be approximately combined with those of seen classes. 
In other words, all the seen classes used in our \shortname{} (Eq. \eqref{eq:srd}) can be viewed {\em collectively} as a basis of the semantic space including unseen classes.

In light of these above considerations,
we further explore the usage of unlabeled data
often available at scale in many real-world situations.
Typically, there is no guarantee that the unlabeled data only contain the seen/target classes
and follow the same distribution as the labeled training set, \ie, {\em unconstrained unlabeled data with unknown distributions and classes}.
This is open-set semi-supervised learning, an emerging problem that has received an increasing amount of attention recently \cite{chen2020semi,yu2020multi,openmatch,T2T}.
%
%
%
Under this interesting context of {\bf \em ``Knowledge distillation meets 
open-set semi-supervised learning''},
we would investigate how knowledge distillation can benefit from unconstrained unlabeled data and previous open-set SSL algorithms, 
as well as how existing open-set SSL methods can influence the knowledge distillation process.


Formally, except the typical labeled training set $\mathcal{D}_l$ with $K$ known classes $\mathcal{Y}$ as used above, we further exploit an unconstrained set $\mathcal{D}_u$ of unlabeled samples not limited to the same label space $\mathcal{Y}$.
This represents a more realistic scenario since unlabeled data is typically collected under little or even no constraints, including the set of class labels considered.
Our objective is to leverage $\mathcal{D}_u$ for further enhancing the student network on top of $\mathcal{D}_l$. 
To that end, we extend the objective function Eq. \eqref{eq:allloss} as:
\begin{equation}\label{eq:combine}
\mathcal{L}_{l+u} = \mathcal{L}_{ce}(\mathcal{D}_{l})+
\alpha \mathcal{L}_{srd}(\mathcal{D}_{l}\cup\mathcal{D}_{u})
+
\beta  \mathcal{R}(\mathcal{D}_{l}\cup\mathcal{D}_{u}),
\end{equation}
where all the loss terms except the cross-entropy $\mathcal{L}_{ce}$ are applied to $\mathcal{D}_u$. We summarize our \shortname{} in Algorithm \ref{alg}.

\vspace{0.1cm}
\noindent{\bf Remarks:}
In the open-set SSL literature, the existing methods \cite{chen2020semi,yu2020multi,openmatch,T2T} 
typically resort to the Out-Of-Distribution (OOD) strategy.
The main idea is to identify and discard those samples not belonging to any seen classes of labeled training set (\ie, OOD samples).
This is driven by a hypothesis that OOD samples are potentially harmful to SSL.
We consider this could be overly restrictive whilst 
ignoring useful knowledge shared across labeled and unlabeled classes, such as common parts and attributes.
For example, flatfish and goldfish exhibit similar body parts such as fins and eyes.
Our \shortname{} and other distillation methods can overcome elegantly this limitation
by leveraging a pretrained teacher model to extract such information for enhancing the training of a student model.
Further, existing open-set SSL works usually consider a small number of  unlabelled data with high similarity as labeled data (\eg, object-centric images sampled from the same source dataset).
In this paper, we scale this setting by using unconstrained unlabeled data at larger scale and with lower similarity (\eg, scene images). 
Under such more realistic settings, 
we reveal new findings in opposite to those reported in previous open-set SSL papers, and show that strategically knowledge distillation is often superior and more reliable than OOD detection in exploiting unconstrained unlabeled data (Sec. \ref{sec:ssl_exp}).

\begin{algorithm}[!htp]
\begin{algorithmic}
\STATE \textbf{Input:} A teacher network $T = \{f^{t},h^{t}\}$, a student network $S = \{f^{s},h^{s}\}$, a labeled dataset $\mathcal{D}_{l}$, an unlabeled dataset $\mathcal{D}_{u}$.  

\STATE \textbf{Output:} Trained $S$.

\STATE \textbf{Per-iteration training process:}\\

\begin{enumerate} \setlength{\itemsep}{-\itemsep}
\item Sampling a mini-batch $\mathcal{B}$ from $\mathcal{D}_{l}$ and  $\mathcal{D}_{u}$;

\item Given an image $I \in \mathcal{B}$,
feeding it through $S$ to obtain the feature vector $\bm{x}^s$ and the logits $\bm{z}^{s}$;

\item Similarly, feeding $I$ through $T$ to obtain the feature vector $\bm{x}^t$ and the logits $\bm{z}^{t}$;

\item Obtaining the cross-network logits (Eq. \eqref{eq:xnet_logit});

\item Computing the objective loss function (Eq. \eqref{eq:combine});

\item Updating $S$ with SGD.

\end{enumerate}
\end{algorithmic}
\caption{\modelname}
\label{alg}
\end{algorithm}

\section{Knowledge Distillation Experiments}
\label{sec:kd_exp}
\subsection{Ablation Study}
\label{sec:ablation}
We first ablate \shortname{} on CIFAR-100 \cite{krizhevsky2009learning}.
For training, we use SGD with weight decay of 5e-4 and momentum of 0.9. 
We set the batch size to 128, the initial learning rate to 0.1 decayed by 0.1 at epochs 100/150 until reaching 200 epochs \cite{heo2019comprehensive}. 
We adopt the standard data augmentation scheme~\cite{zagoruyko2016paying} including random cropping (w/ 4-pixels padding) and horizontal flipping.
By default, we utilize two variants of Wide ResNet \cite{wrn}, namely WRN40-4 and WRN16-4, as the teacher and student, unless specified otherwise.

\vspace{0.1cm}
\noindent\textbf{\shortname{} loss designs:} 
We first evaluate the three different designs of \shortname{} loss discussed in Sec.~\ref{sec:method_SRD}.
The corresponding experiments are shown in Table~\ref{tab:ab_srd} that all these designs are effective with $\mathcal{L}_{srd}^{mse}$ (Eq. \eqref{eq:srd_mse}) yielding the best results. Interestingly, $\mathcal{L}_{srd}^{mse}$ even slightly surpasses the teacher's performance.
We hypothesize this is due to that our \shortname{} might impose some regularization effect (\eg, fusing the capacity of the student and teacher to some degree) during distillation.
Overall, this validates the efficacy of our \shortname{} formulation and loss design.
In the following experiments, we hence use 
$\mathcal{L}_{srd}^{mse}$ as the default design, unless stated otherwise.

\begin{table}[h]
\caption{Evaluation of different \shortname{} loss designs on CIFAR-100. 
Teacher: WRN40-4;
Student: WRN16-4.
}
\centering
\begin{tabular}{l|cc}
\hline
Accuracy&Top-1 (\%)&Top-5 (\%)\\
\hline
\em Supervised learning &76.97&93.89\\
\hline
$\mathcal{L}_{srd}^{kl}$ (Eq. \eqref{eq:srd_kl})&79.04&95.12\\\\[-5pt]
$\mathcal{L}_{srd}^{mse}$ (Eq. \eqref{eq:srd_mse})&\textbf{79.58}&\textbf{95.21}\\\\[-5pt]
$\mathcal{L}_{srd}^{pmse}$ (Eq. \eqref{eq:srd_msev2})&79.13&94.88 \\
\hline
Teacher &79.50&94.57\\
\hline
\end{tabular}
\label{tab:ab_srd}
\vspace{-5mm}
\end{table}

\vspace{0.1cm}
\noindent\textbf{Effect of loss components:}
We examine the impact of 
{{\color{red} each loss component} in Eq.~\eqref{eq:allloss}.
\revision{As shown in Table \ref{tab:ab_loss}, around $1\%$ and $2\%$ improvements in Top-1 accuracy can be obtained by the regularization and our distillation, respectively.
So, our SRD loss is clearly more effective than $\mathcal{R}$.
Moreover, when combining them together, an additional $0.48\%$ improvement is gained.}

\begin{table}[h]
\centering
\caption{Effect of loss components on CIFAR-100.}

\begin{tabular}{c|c|c|c}
\hline
$\mathcal{R}$ &$\mathcal{L}_{srd}$ & Top-1 (\%) &Top-5 (\%)\\
\hline
\xmark &\xmark &76.97&93.89\\
\cmark &\xmark &78.05&94.45\\
\xmark &\cmark &79.10&94.99 \\
\cmark &\cmark &\textbf{79.58} &\textbf{95.21}\\
\hline
\end{tabular}
\label{tab:ab_loss}
\vspace{-5mm}
\end{table}

\vspace{0.1cm}
\noindent\textbf{Distillation effect:}
The objective of knowledge distillation is to encourage the student mimic the prediction behaviour of the teacher.
It is hence insightful to evaluate this mimicry quality.
For comparative evaluation, we contrast \shortname{}
with the logit-matching distillation \cite{hinton2015distilling}.
For mimicry measurement, we adopt the KL divergence between the teacher's and student's predictions, as well as $L_2$ distance between the teacher and student representations.
It is observed from Table~\ref{tab:ans_sim} that the mimicry ability of a student presents a positive correlation with its accuracy.
More similarly a student can mimic the teacher, more accurate result achieved.
Further, we examine qualitatively the feature representation distribution.
It is evident in Fig. \ref{fig:tsne} that our \shortname{} can learn more discriminative features, consistent with the above numerical measurement.
Besides, Top-1 accuracy with pretrained teacher classifier proves that learned representation by \shortname{} is closer to teacher's feature.

\vspace{0.1cm}
\noindent{\bf Prediction confidence:}
\revision{We examine the distribution of prediction confidence across true positives (TP), false positives (FP), false negatives (FN), and true negatives (TN).
As shown in Fig. \ref{fig:conf}, the peak confidence levels for TP and TN correspond to correct predictions, whereas for FP and FN, they are associated with incorrect predictions. Compared to KD, SRD has higher confidence for correct predictions and low confidence for incorrect predictions, which is a favored property.}


\begin{table}[h]
\centering
\caption{
Evaluating the distillation effect (\ie, mimicry quality) on CIFAR-100. Metrics: KL divergence between the student's and teacher's predictions, $L_2$ distance between the student's and teacher's representations, Top-1 accuracy with teacher's classifier, and Top-1 accuracy with individual classifier.}
\resizebox{1\linewidth}{!}{
\setlength\tabcolsep{3.5pt}
\begin{tabular}{l|c|c|c|c}
\hline
Method & \tabincell{c}{KL div.}& \tabincell{c}{$L_2$ dis.}
&\tabincell{c}{with $\mathbf{W}^t$}
&\tabincell{c}{Top-1 (\%)}\\
\hline
\em Supervised learning &0.5964 &1.48 &0.91 &76.97 \\ \hline
KD~\cite{hinton2015distilling} & 0.5818 &1.21 &1.15& 78.35\\
\hline
\bf \shortname &\textbf{0.4597}&\textbf{1.01}&\textbf{79.11}&\textbf{79.58}\\
\hline
\end{tabular}}
\label{tab:ans_sim}
\end{table}

\begin{figure}[h]
\centering
\subfigure[{\fontsize{6}{7}\selectfont Supervised learning}]{\includegraphics[height=0.148\textwidth]{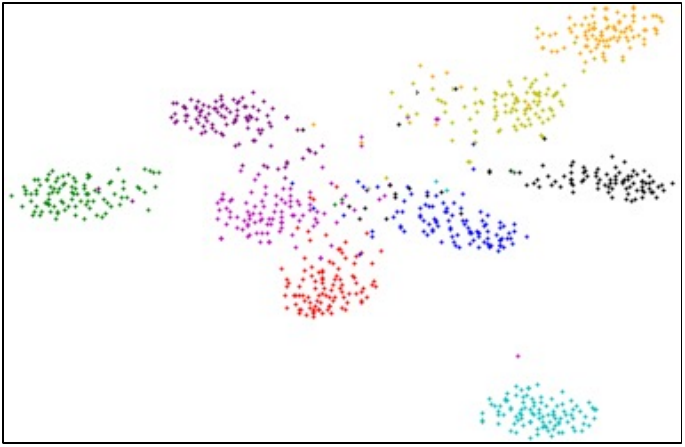}}
\subfigure[{\fontsize{6}{7}\selectfont Teacher}]{\includegraphics[height=0.148\textwidth]{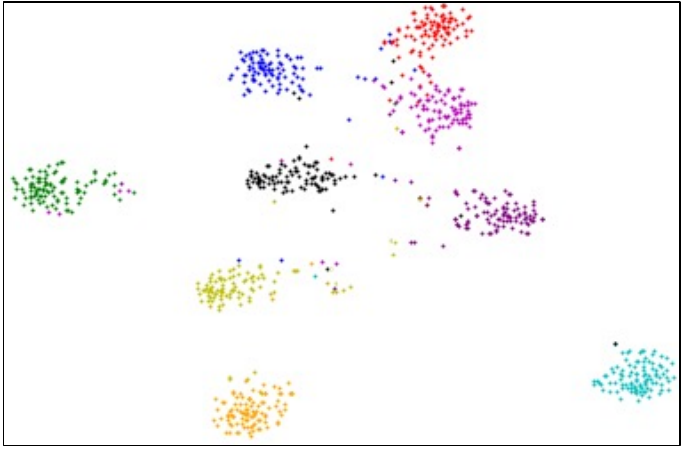}}
\subfigure[{\fontsize{6}{7}\selectfont KD}]{\includegraphics[height=0.148\textwidth]{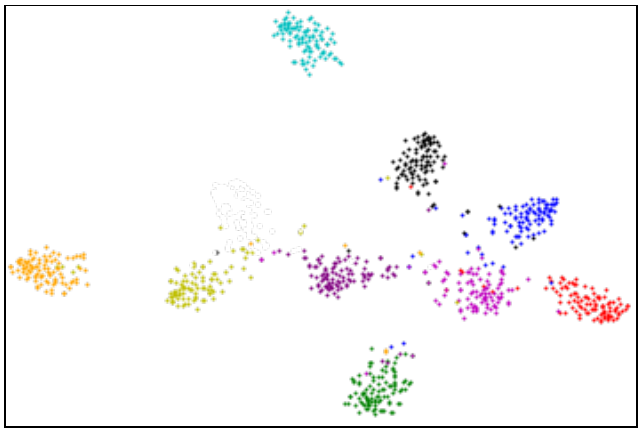}}
\subfigure[{\fontsize{6}{7}\selectfont \shortname}]{\includegraphics[height=0.148\textwidth]{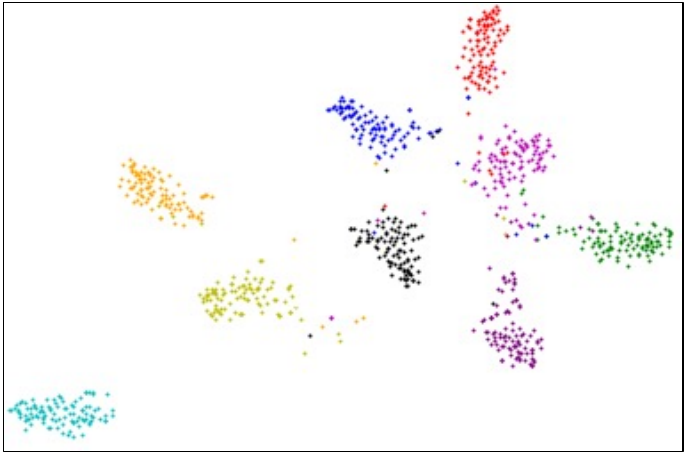}}
\caption{Feature distribution visualization of 10 classes on 
CIFAR-100. Class is color coded.
Better viewed in color.
}
\label{fig:tsne}
\end{figure}

\begin{figure}[h]
\centering
\subfigure[{\fontsize{6}{7}\selectfont TP}]{\includegraphics[height=0.148\textwidth]{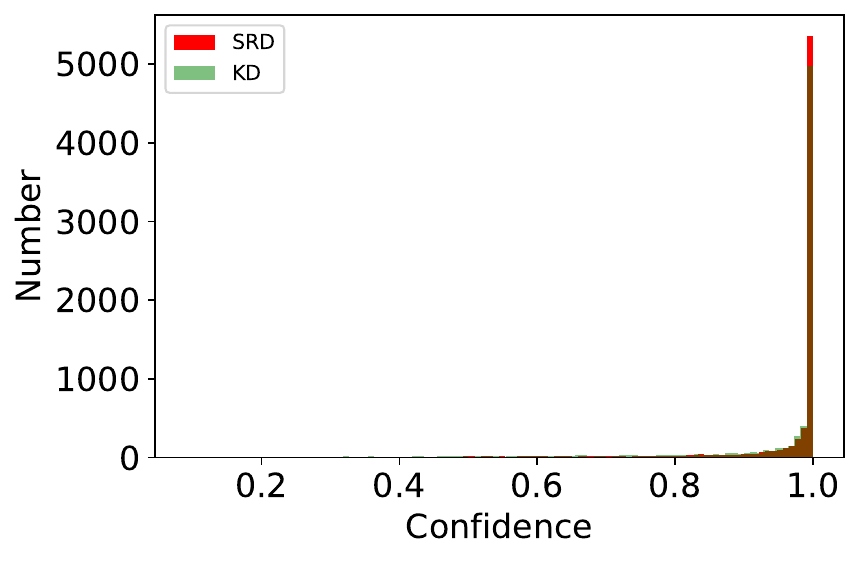}}
\subfigure[{\fontsize{6}{7}\selectfont FP}]{\includegraphics[height=0.148\textwidth]{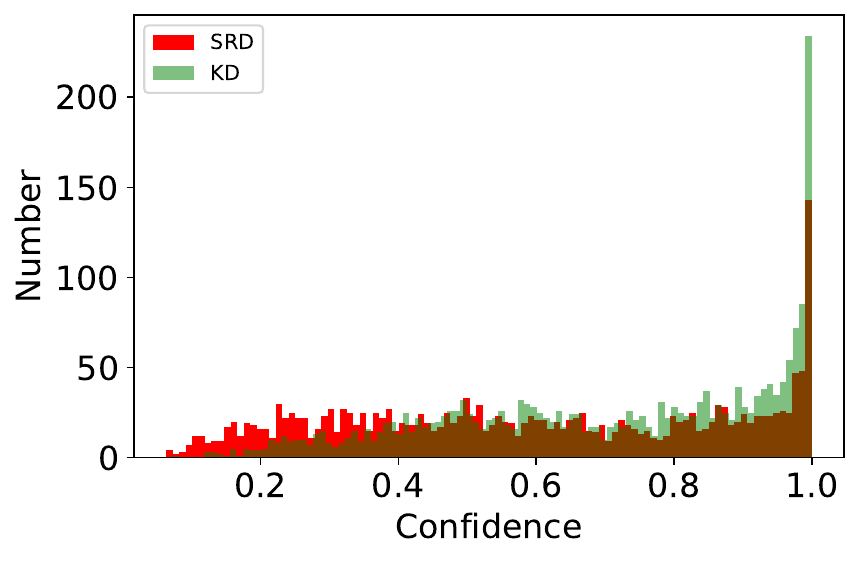}}
\subfigure[{\fontsize{6}{7}\selectfont FN}]{\includegraphics[height=0.148\textwidth]{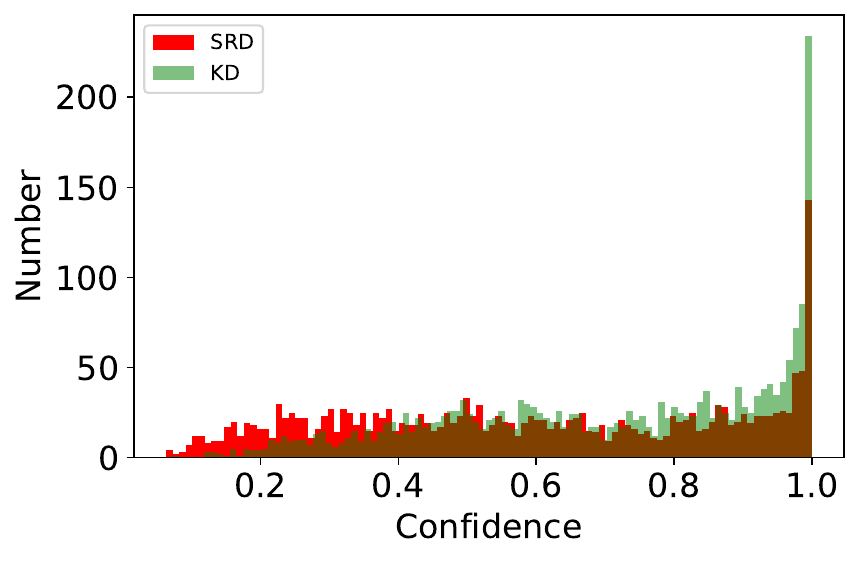}}
\subfigure[{\fontsize{6}{7}\selectfont TN}]{\includegraphics[height=0.148\textwidth]{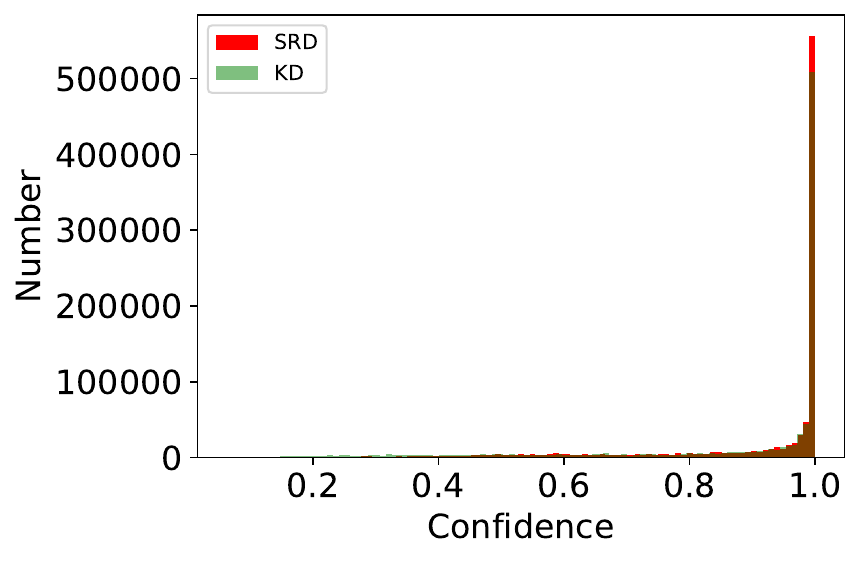}}
\caption{\added{Classifier confidence distribution over TP, FP, TN, FN by SRD and KD on CIFAR-100 test set.
}}
\label{fig:conf}
\vspace{-4mm}
\end{figure}


\vspace{0.1cm}
\noindent\textbf{Complementary with logit-matching distillation:}
We further test the complementary of our \shortname{} with 
the logit-matching distillation \cite{hinton2015distilling}.
We optimize their combination weight by typical grid search.
For extensive evaluation, we experiment with a diverse set of teacher/student pairs using ResNets~\cite{He_2016_CVPR}, WRNs~\cite{zagoruyko2016wide}, MobileNetV2~\cite{mobilenetv2}.
Table~\ref{tab:sup_c100_c} shows that
this distillation combination is compatible and often leads to further
performance gain.

\begin{table}[ht]
\centering

\caption{Complementary with logit-matching knowledge distillation \cite{hinton2015distilling} on CIFAR-100.
Metric: Top-1 accuracy ($\%$).
Surp. learn.: Supervised learning.
}
\resizebox{1\linewidth}{!}{
\setlength\tabcolsep{3.5pt}
\begin{tabular}{l|c|c|c|c|c|c}
\hline
Teacher     &WRN40-4&WRN40-4&ResNet34&ResNet50&ResNet34&WRN40-4\\
(Params)    &8.97M&8.97M&1.39M&1.99M&21.33M&8.97M\\
\hline
Student     &WRN16-2&WRN16-4&ResNet10&ResNet18&WRN16-2&MobileNetV2\\
(Params)    &0.70M &2.77M&0.34M&0.75M&0.70M&2.37M\\
\hline
\em Surp. learn.    &72.70&76.97&68.42&71.07&72.70&68.42\\
\hline
\bf \shortname &75.96&79.58&69.91&73.47&75.38&71.82\\
{\bf \shortname}+KD\cite{hinton2015distilling} &\bf{76.05}&\bf{79.63}&\bf{70.41}&\bf{73.53}&\bf{75.46}&\bf{72.06}\\
\hline
Teacher     &79.50&79.50&72.05&73.31&78.44&79.50\\
\hline
\end{tabular}}
\label{tab:sup_c100_c}
\end{table}

\vspace{0.1cm}
\noindent\textbf{Evaluating the generality of architecture:}
Except CNNs as the backbone, we further adopt the recent Vision Transformers (ViTs) \cite{dosovitskiy2020image} to evaluate the architectural generality of our \shortname{}.
We experiment with two teacher-student pairs: ViT-Base as the teacher and ViT-Tiny as the student, and ResNet50 as teacher and ViT-Tiny as student (cross-architecture).
We use the popular code repository\footnote{\url{https://github.com/rwightman/pytorch-image-models}}.
In Table \ref{tab:vit} we observe that our \shortname{} is again superior over top alternatives, consistent with the case of pure CNN backbones.
This suggests our method is architecture agnostic.

\begin{table}[ht]
\centering
\caption{Architectural generality evaluation on CIFAR-100. Metric: Top-1 accuracy ($\%$).
}
\resizebox{1\linewidth}{!}{
\setlength\tabcolsep{3.5pt}
\begin{tabular}{c|c|c}
\hline
Teacher & ViT\_Base (85.55M) &ResNet50(23.73M)\\
Student & ViT\_Tiny (5.54M)  &ViT\_Tiny (5.54M)\\\hline
\em Supervised learning& 80.43 & 80.43\\\hline
KD~\cite{hinton2015distilling}&82.11&81.40\\
CRD~\cite{tian2019contrastive}&83.62&82.28\\
\bf \shortname& \bf{85.34}  &\bf{83.37}\\\hline
Teacher & 93.86 &85.89\\\hline
\end{tabular}}
\label{tab:vit}
\end{table}

\subsection{Comparison with State-of-the-Art Methods} \label{sec:sup_sota}


\noindent {\bf Architectures:}
For extensive evaluation, we consider multiple mainstream network architectures including 
ResNets~\cite{He_2016_CVPR}, Wide ResNets~\cite{zagoruyko2016wide}, MobileNetV2~\cite{mobilenetv2}, and MobileNet~\cite{howard2017mobilenets} with different learning capacities.


\vspace{0.1cm}
\noindent {\bf Competitors:}
We compare our \shortname{} with five state-of-the-art knowledge distillation methods: KD~\cite{hinton2015distilling}, AT~\cite{zagoruyko2016paying}, 
OFD~\cite{heo2019comprehensive}, RKD~\cite{park2019relational}, and CRD~\cite{tian2019contrastive}.

\vspace{0.1cm}
\noindent {\bf Application tasks:}
Except the common image classification problem, 
we further consider two more practically critical applications with less investigation in distillation:
fine-grained face recognition (Sec. \ref{sec:exp_face}), and
binary network optimization (Sec. \ref{sec:exp_bin_net}).


\subsubsection{Evaluation on CIFAR-10} 
\label{ssec:CIFAR-10}
\noindent\textbf{Setting:}
CIFAR-10 is a popular image classification dataset consisting of 50,000 training and 10,000 testing images evenly distributed across 10 object classes. 
All the images have a resolution of $32\times 32$ in pixel. Following~\cite{zagoruyko2016paying}, during training, we randomly crop and horizontally flip each image. 
We train the ResNet for 350 epochs using SGD. We set the initial learning rate to 0.1, gradually reduced by a factor of 10 at epochs 150, 250 and 320. Similarly, we train the WRN models for 200 epochs with the initial learning rate of 0.1 and a decay rate of 5 at epochs 60, 120 and 160. We set the dropout rate to 0. 
For the logit-matching KD~\cite{hinton2015distilling}, we set $\alpha=0.9$ and $T=4$. For AT~\cite{zagoruyko2016paying}, as in~\cite{zagoruyko2016paying,tung2019similarity}, we set the weight of distillation loss to 1000. 
Note the AT loss is added after each layer group for WRN and the last two groups for ResNet following~\cite{zagoruyko2016paying}.
Following OFD~\cite{heo2019comprehensive}, we set the weight of distillation loss to $10^{-3}$.
For RKD~\cite{park2019relational}, we set $\beta_{1} = 25$ for distance, and $\beta_{2} = 50$ for angle, as suggested in~\cite{park2019relational,tian2019contrastive}.
We exclude CRD~\cite{tian2019contrastive} here because, in our experiments, we found that the parameters originally proposed for CIFAR-100 and ImageNet-1K do not work well for CIFAR-10. 
We evaluate three types of teacher/student pairs:
(1) Two pairs using WRNs; (2) Three pairs using ResNets;
(3) One pair across WRN and ResNet. 

\begin{table}[ht]
\centering
\caption{Evaluating knowledge distillation methods on CIFAR-10.
Metric: Top-1 accuracy ($\%$).
The parameter size of each network is given in the round bracket.
Surp. learn.: Supervised learning.
}
\resizebox{1\linewidth}{!}{
\setlength\tabcolsep{3.5pt}
\begin{tabular}{l|c|c|c|c|c|c}
\hline
Teacher     &WRN16-2&WRN40-2&ResNet26&ResNet26&ResNet34&ResNet26\\
(Params)&(0.69M) &(2.2M)  &(0.37M) &(0.37M) &(1.4M)  &(0.37M)\\
\hline
Student     &WRN16-1 &WRN16-2 &ResNet8&ResNet14&ResNet18&WRN16-1\\
(Params) &(0.18M) &(0.69M) &(0.08M) &(0.17M) &(0.7M) &(0.18M)\\
\hline
\em Surp. learn. &91.04 &93.98&87.78&91.59&93.35&91.04\\
\hline
KD~\cite{hinton2015distilling}&92.57&94.46&88.75&92.57&93.74&92.42\\
AT~\cite{zagoruyko2016paying}&92.15&94.39&88.15&92.11&93.52&91.32\\
OFD~\cite{heo2019comprehensive}&92.28&94.30&87.49&92.51&93.80&92.47\\
RKD~\cite{park2019relational}&92.51&94.41&88.50&92.36&92.95&92.08\\
\bf \shortname&\bf{92.95}&\bf{94.66}&\bf{89.02}&\bf{92.70}&\bf{93.92}&\bf{92.94}\\
\hline
Teacher   &93.98 &95.07 &93.58 &93.58 &94.11 &93.58\\
\hline
\end{tabular}}
\label{tab:sup_c10}
\vspace{-4mm}
\end{table}

\vspace{0.1cm}
\noindent\textbf{Results:}
Top-1 classification results on CIFAR-10 are compared in Table~\ref{tab:sup_c10}. 
On this largely saturated dataset, it is evident that our \shortname{} still results in clear gains over all the alternatives in all settings, suggesting a generic and stable superiority. 
Besides, it is found that the logit-matching KD performs second only to \shortname{}, surpasses all the other KD variants.

\subsubsection{Evaluation on CIFAR-100} 
\label{ssec:CIFAR-100}
\noindent\textbf{Setting:}
We use the same setting as in ablation (Sec. \ref{sec:ablation}).
Similarly, we experiment with three sets of teacher/student network pairs. 
The {\em first} set is constructed by the performance using WRNs:
strong teacher/weak student (WRN40-4/WRN16-2), and
strong teacher/fair student (WRN40-4/WRN16-4).
The {\em second} set repeats similar pairings using ResNets:
(ResNet34/ResNet10), and (ResNet50/ResNet18).
The {\em third} set is created by combining different architectural families:
(ResNet34/WRN16-2) and (WRN40-4/MobileNetV2).

\begin{table}[ht]
\centering
\caption{Evaluating knowledge distillation methods on CIFAR-100.
Metric: Top-1 accuracy ($\%$).
Surp. learn.: Supervised learning.
}
\resizebox{1\linewidth}{!}{
\setlength\tabcolsep{3.5pt}
\begin{tabular}{l|c|c|c|c|c|c}
\hline
Teacher     &WRN40-4&WRN40-4&ResNet34&ResNet50&ResNet34&WRN40-4\\
(Params)    &(8.97M)&(8.97M)&(1.39M)&(1.99M)&(21.33M)&(8.97M)\\
\hline
Student     &WRN16-2&WRN16-4&ResNet10&ResNet18&WRN16-2&MobileNetV2\\
(Params)    &(0.70M) &(2.77M)&(0.34M)&(0.75M)&(0.70M)&(2.37M)\\
\hline
\em Surp. learn. &72.70&76.97&68.42&71.07&72.70&68.42\\
\hline
KD~\cite{hinton2015distilling} &74.52&78.35&69.18&73.41&73.95&69.15\\
AT~\cite{zagoruyko2016paying}  &74.33&78.06&68.49&71.90&72.32&68.95\\
OFD~\cite{heo2019comprehensive}&75.57&79.29&68.94&72.79&74.78&70.08\\
RKD~\cite{park2019relational}  &74.23&78.38&68.70&70.93&73.91&68.19\\
CRD~\cite{tian2019contrastive} &75.27&78.83&\bf{70.24}&73.23&74.88&71.46\\
\bf \shortname                 &\bf{75.96}&\bf{79.58}&69.91&\bf{73.47}&\bf{75.38}&\bf{71.82}\\
\hline
Teacher     &79.50&79.50&72.05&73.31&78.44&79.50\\
\hline
\end{tabular}}
\label{tab:sup_c100}
\vspace{-4mm}
\end{table}

\noindent\textbf{Results:}
We report the Top-1 performance on CIFAR-100 in Table~\ref{tab:sup_c100}. 
On this more challenging test, we observe that for almost all teacher/student configurations, our \shortname{} achieves consistent and significant accuracy gains over prior methods. Further, there is no clear second best. For WRN pairs, OFD ranks second. For the other cases, CRD instead achieves the second.
This suggests that our \shortname{} is also more generalizable to
different distillation configurations.


\subsubsection{Evaluation on ImageNet-1K}
\noindent\textbf{Setting:}
For larger scale evaluation, we test the standard ImageNet-1K benchmark. 
We crop the images to a resolution of $224\times 224$ pixels for both training and test. 
We use SGD with Nesterov momentum set to 0.9, weight decay to $10^{-4}$, initial learning rate to 0.2 which decays by a factor of 10 every 30 epochs.
We set the batch size to 512. 
We train a total of 100 epochs for all methods except CRD \cite{tian2019contrastive} which uses 10 more epochs following the authors' suggestion. 
For simplicity, we use pretrained PyTorch models~\cite{paszke2017automatic} as the teacher ~\cite{heo2019comprehensive,tian2019contrastive}.
We adopt two common teacher/student pairs:
ResNet34/ResNet18 and ResNet50/MobileNet~\cite{howard2017mobilenets}.
When testing logit-matching KD \cite{hinton2015distilling}, we set the weight of KL loss and cross-entropy loss to 0.9 and 0.5, which yields better accuracy as found in \cite{tian2019contrastive}. 


\vspace{0.1cm}
\noindent\textbf{Results:}
We report the ImageNet classification results in Table \ref{tab:imagenet}.
Again, we observe that our \shortname{} outperforms all the competitors
by a large margin in all cases.
This suggests that the advantage of \shortname{} is scalable.
Specifically, for the ResNet34/ResNet18 pair, RKD reaches the second best Top-1 accuracy; Whilst in the ResNet-50/MobileNet case, CRD is the second.
This further suggests that previous methods are less stable than \shortname{} in the selection of networks.
Critically, \shortname{} favors MobileNet with 
a big margin of absolute $1.1\%$ in absolute terms over the best alternative, CRD.
Considering that MobileNet has been widely used 
across many devices and mobile platforms, this performance gain
could be particularly promising and valuable in practice.



\begin{table}[!htbp]
\centering
\caption{Evaluating knowledge distillation methods on ImageNet-1K.
Metric: Top-1 and Top-5 accuracy ($\%$).
The parameter size of each network is given in the round bracket.
Surp. learn.: Supervised learning.
}
\resizebox{1\linewidth}{!}{
\setlength\tabcolsep{3.5pt}
\begin{tabular}{l|c|c|c|c}
\hline
Teacher &\multicolumn{2}{c}{ResNet34 (21.80M)} &\multicolumn{2}{c}{ResNet50 (25.56M)}\\
Student &\multicolumn{2}{c}{ResNet18(11.69M)}&\multicolumn{2}{c}{MobileNet(4.23M)}\\
\hline
&Top-1 (\%)&Top-5 (\%)&Top-1 (\%)&Top-5 (\%)\\
\hline
\em Surp. learn. &70.04 &89.48 &70.13&89.49\\
\hline
KD~\cite{hinton2015distilling} &70.68 &90.16 &70.68 &90.30\\
AT~\cite{zagoruyko2016paying}  &70.59 &89.73 &70.72 &90.03\\
OFD~\cite{heo2019comprehensive}&71.08 &90.07 &71.25 &90.34\\
RKD~\cite{park2019relational}  &71.34 &90.37 &71.32 &90.62\\
CRD~\cite{tian2019contrastive} &71.17 &90.13 &71.40 &90.42\\
\bf \shortname                 &\bf{71.73} &\bf{90.60} &\bf{72.49}&\bf{90.92}\\
\hline
Teacher &73.31 &91.42 &76.16&92.86\\
\hline
\end{tabular}}
\label{tab:imagenet}
\vspace{-4mm}
\end{table}

\subsubsection{Evaluation on face recognition}
\label{sec:exp_face}

Except coarse object classification,
we also consider fine-grained face recognition task
that requires learning more detailed representations specific to individual object instance in the distillation perspective.

\vspace{0.1cm}
\noindent {\bf Datasets:}
For training, we use the MS1MV2 dataset \cite{arcface}, a refined version of MS-Celeb-1M \cite{guo2016ms,wu2018light}.
For testing, we use the refined MegaFace~\cite{guo2016ms,kemelmacher2016megaface} which includes a million of distractors. 
We adopt MegaFace's Challenge1 using FaceScrub as the probe set.

\vspace{0.1cm}
\noindent {\bf Performance evaluation:}
We consider two face recognition tasks.
(1) {\em Face verification}: Given a pair of face images, the objective is to determine whether they describe the same person's identity.
This is accomplished by calculating a pairwise similarity (\eg, cosine similarity or negative Euclidean distance) in the feature space and matching it \wrt{} a threshold.
For performance metrics, we adopt the True Acceptance Rate (TAR) at the False Acceptance Rate (FAR) of $10^{-6}$ \cite{arcface}. 
The decision threshold is determined by the FAR.
(2) {\em Face identification}: Given a query face image, the objective is to identify those images with the same person identity in a gallery.
This is often treated as a retrieval problem
by ranking the gallery images according to the pairwise similarity scores \wrt{} the query image.
To evaluate the performance, we use the rank-1 accuracy \cite{arcface}.


\vspace{0.1cm}
\noindent\textbf{Competitors:} 
For comparative evaluation, we consider both logit (KD~\cite{hinton2015distilling}) and feature (AT~\cite{zagoruyko2016paying}, RKD~\cite{park2019relational} and PKT~\cite{passalis2018learning}) based distillation methods. 
Note, here we exclude CRD~\cite{tian2019contrastive} due to its optimization difficulty with margin-based softmax loss
and OFD~\cite{heo2019comprehensive} due to its difficulty of finding a good-performing distillation layer.


\vspace{0.1cm}
\noindent\textbf{Setting:} 
For model training, we use SGD as the optimizer and set the momentum to 0.9, and weight decay to $5e-4$. We set the batch size to 512.
We set the initial learning rate to 0.1, and decay it by 0.1 at 100K, and 160K iterations.
We train each model for a total of 180K iterations. 
We adopt ResNet101 \cite{He_2016_CVPR} as the teacher with 65.12M parameters, and MobileFaceNet \cite{chen2018mobilefacenets} as the student with 1.2M parameters.


\vspace{0.1cm}
\noindent
\textbf{Results:} We present the face recognition results in Table~\ref{tab:megaface}.
We make several key observations as follows.
{\bf (1)} Our \shortname{} achieves the best distillation in comparison to all the other competing methods, suggesting its superior ability of distilling fine-grained representational knowledge.
{\bf (2)} For face verification, \added{only SRD and AT gain advantages from distillation, with SRD outperforming all its rivals. Several reasons explain this performance observation: First, there is a pronounced discrepancy in performance levels between the teacher and student models, adding difficulty in distillation \cite{wang2021knowledge}. Second, while existing methods focus on guiding the student's learning with specifically designed knowledge at either the intermediate or classification stages, SRD uniquely distills knowledge directly from the final representation which is used directly for face verification. Thirdly, previous methods often ignore the class prototypes within the teacher's classifier, which capture the discrimination information for each identity. In contrast, our strategy makes full use of the teacher's classifier by feeding the student's features into the teacher's classifier.}
{\bf (3)} The logit-matching KD~\cite{hinton2015distilling} fails on both tasks. 
A plausible reason is due to incompatibility between softened logits matching and margin-based softmax loss. By leveraging the learned identity prototypes with teacher's classifier to constrain the learning of student's representation, \shortname{} manages to distil useful knowledge with subtlety successfully.

\begin{table}[!htbp]
\centering
\caption{Face identification and verification results on MegaFace.
Teacher: ResNet101; Student: MobileFaceNet.}
\begin{tabular}{l|c|c}
\hline
Method                        &Verification (\%)  &Identification (\%)\\
\hline
\hline
\em Supervised learning                        &93.44    &92.28 \\
\hline
KD~\cite{hinton2015distilling} &92.86    &90.91\\
AT~\cite{zagoruyko2016paying}  &93.55    &92.46\\
RKD~\cite{park2019relational}  &93.37    &92.34\\
PKT~\cite{passalis2018learning}&93.25    &92.38\\
\bf \shortname                 &\bf{94.17}    &\bf{93.26}\\
\hline
Teacher                        &98.56    &98.82\\ 
\hline
\end{tabular}
\label{tab:megaface}
\vspace{-4mm}
\end{table}

\subsubsection{Evaluation on binary network distillation}
\label{sec:exp_bin_net}

In all the above experiments, both student and teacher use some networks
parameterized with real-valued precision (real networks).
However, they are often less affordable in low-resource regime (\eg, mobile devices).
One of the promising approaches is to deploy neural networks
with binary-valued parameters (\ie, binary networks, the most extreme case of network quantization), as they are not only smaller in size but run faster and more efficiently~\cite{rastegari2016xnor,bulat2019xnor}.
However, training accurate binary networks from scratch is highly challenging and the use of distillation has been shown to be a key component~\cite{martinez2020training}.
In light of these observations, we investigate the largely ignored yet practically critical binary network distillation problem.
The objective is to distil knowledge from a real teacher to a binary student. 


\vspace{0.1cm}
\noindent\textbf{Datasets:} 
In this evaluation, we use CIFAR-100 and ImageNet-1K
following the standard setup as above.

\vspace{0.1cm}
\noindent\textbf{Competitors:} 
Similar as face recognition, we compare with both logit (KD~\cite{hinton2015distilling}) and feature (AT~\cite{zagoruyko2016paying}, RKD~\cite{park2019relational}, OFD~\cite{heo2019comprehensive}, PKT~\cite{passalis2018learning}, and CRD~\cite{tian2019contrastive}) based distillation methods.

\vspace{0.1cm}
\noindent\textbf{Setting:}
For training, we use Adam as the optimizer.
For CIFAR-100, we set the initial learning 0.001 decayed by a factor of 0.1 at the epochs of $\{150,250,320\}$ and the total epochs to 350.
For ImageNet-1K, we use the initial learning 0.002 decayed by a factor of 0.1 at the epochs of $\{30,60,90\}$, and a total of 100 epochs. 
For the teacher and student, we use the same ResNet architecture with the modifications as introduced in ~\cite{bulat2019xnor}.

\vspace{0.1cm}
\noindent\textbf{Results:}
We report binary network distillation results in Table~\ref{tab:bin}.
It is evident that on both datasets \shortname{} outperforms consistently all the alternatives by a clear margin.
On the other hand, OFD \cite{heo2019comprehensive} performs worst probably due to its problem specific and less generalizable distillation position.
Overall, these results validate that the performance advantage of \shortname{} can extend well from real networks to binary networks, a rarely investigated but practically significant application scenario.

\begin{table}[h]
\centering
\caption{Binary network distillation results on CIFAR-100 and ImageNet-1K.
Metric: Top-1 accuracy (\%).
Surp. learn.: Supervised learning.}
%
\begin{tabular}{l|c|c}
\hline
Datasets &CIFAR-100 &ImageNet-1K\\
\hline
\hline
Teacher &ResNet34(Real) &ResNet18(Real)\\
Student &ResNet34(Binary) &ResNet18(Binary)\\
\hline
\em Surp. learn. &65.34    &56.70 \\
\hline
KD~\cite{hinton2015distilling} &68.65    &57.39\\
AT~\cite{zagoruyko2016paying}  &68.54    &58.45\\
OFD~\cite{heo2019comprehensive}&66.84    &55.74\\
RKD~\cite{park2019relational}  &68.61    &58.84\\
CRD~\cite{tian2019contrastive} &68.78    &58.25\\
\bf \shortname                 &\bf{70.50}    &\bf{59.57}\\
\hline
Teacher &75.08    &70.20\\
\hline
\end{tabular}
\label{tab:bin}
\vspace{-4mm}
\end{table}


\section{Open-Set Semi-Supervised Experiments}
\label{sec:ssl_exp}

\noindent{\bf Datasets: } 
In this evaluation,
we use {\em CIFAR-100}~\cite{krizhevsky2009learning}
as the target dataset, including a labeled training set and a test set. 
We follow the standard training-test split.
To simulate a realistic open-set SSL setting,
we use {\em Tiny-ImageNet}~\cite{le2015tiny} as unlabeled training data.
As a subset of ImageNet-1K~\cite{ILSVRC15},
this dataset consists of 200 classes, each with 500, 50 and 50 images for training, validation and test, respectively.
We only use its training set, consisting of 100,000 images.
The two datasets share a very small proportion of classes.

\vspace{0.1cm}
\noindent{\bf Implementation details: }
To facilitate fair comparative evaluation,
we adopt the same setup of~\cite{tian2019contrastive}
including the training configuration as given in its open source code\footnote{\url{https://github.com/HobbitLong/RepDistiller}}.
For all compared methods below, the same initialization,
training and test data are applied under the same training setup.
We apply the same data augmentation for all the labeled and unlabeled training data as in Sec.~\ref{ssec:CIFAR-100}.
We resize all the images to $32 \times 32$ before data augmentations.

\subsection{Evaluation on Knowledge Distillation Methods}
\label{sec:unsup_kd}

\noindent {\bf Setting: }
We consider a diverse set of three (teacher, student) network pairs for different distillation methods:
(ResNet32$\times$4, ResNet8$\times$4),
(WRN40-2, WRN40-1), and (ResNet32x4, ShuffleNetV1).
For statistical stability, for each experiment we run 5 trials and report the average result.


\vspace{0.1cm}
\noindent {\bf Competitors: }
We compare extensively a total of 12 state-of-the-art distillation methods:
KD~\cite{hinton2015distilling}, 
FitNet~\cite{romero2015fitnets},
AT~\cite{zagoruyko2016paying},
SP~\cite{tung2019similarity},
CC~\cite{peng2019correlation},
VID~\cite{ahn2019variational}, 
RKD~\cite{park2019relational}, 
PKT~\cite{passalis2018learning},
AB~\cite{heo2019knowledge},
FT~\cite{kim2018paraphrasing},
NSP~\cite{huang2017like},
CRD~\cite{tian2019contrastive}.
For all these methods except CRD~\cite{tian2019contrastive}, 
unlabeled data can be directly accommodated without design adaptation.
Instead, class labels are required by CRD \cite{tian2019contrastive}.
To solve this issue, we extend CRD by a pseudo-labeling strategy.
We obtain a pseudo label for every unlabeled sample in a maximum likelihood manner using the teacher network pretrained on the labeled set.
We then treat the pseudo labels as the ground-truth during training CRD.

\begin{table*}[ht]
\centering
\caption{Comparing distillation methods under the open-set semi-supervised learning setting on CIFAR-100.
Labeled training data $\mathcal{D}$: CIFAR-100;
Unlabeled training data $\mathcal{U}$: Tiny-ImageNet.
Metric: Top-1 accuracy (\%).
The parameter size of each network is given in the round bracket. 
Surp. learn.: Supervised learning.
}

\begin{tabular}{l|cc|cc|cc}
\hline
Teacher &ResNet$32\times4$&(7.43M)   &WRN40-2&(2.25M) &ResNet$32\times4$&(7.43M)\\
Student &ResNet$8\times4$&(1.23M)    &WRN40-1&(0.57M) &ShuffleNetV1&(0.94M)\\
\hline
\hline
Training data &$\mathcal{D}$ &$\mathcal{D+U}$
&$\mathcal{D}$ &$\mathcal{D+U}$
&$\mathcal{D}$ &$\mathcal{D+U}$\\
\hline
\em Surp. learn. &72.50 &- &71.98&- &70.59&-\\
\hline
KD~\cite{hinton2015distilling} &73.33$\pm$0.25 &74.68$\pm$0.05\textcolor{green}{$\uparrow$} &73.54$\pm$0.20 &75.08$\pm$0.25\textcolor{green}{$\uparrow$} &74.07$\pm$0.19&76.52$\pm$0.03\textcolor{green}{$\uparrow$}\\
FitNet~\cite{romero2015fitnets}&73.50$\pm$0.28 &73.30$\pm$0.14\textcolor{red}{$\downarrow$} &72.24$\pm$0.24&71.43$\pm$0.17\textcolor{red}{$\downarrow$}
&73.59$\pm$0.15&72.83$\pm$0.13\textcolor{red}{$\downarrow$}\\
AT~\cite{zagoruyko2016paying}  &73.44$\pm$0.19 &71.75$\pm$0.11\textcolor{red}{$\downarrow$} &72.77$\pm$0.10&73.11$\pm$0.19\textcolor{green}{$\uparrow$}
&71.73$\pm$0.31&72.82$\pm$0.24\textcolor{green}{$\uparrow$}\\
SP~\cite{tung2019similarity}   &72.94$\pm$0.23 &72.10$\pm$0.21\textcolor{red}{$\downarrow$} &72.43$\pm$0.27&73.02$\pm$0.13\textcolor{green}{$\uparrow$} &73.48$\pm$0.42&76.01$\pm$0.15\textcolor{green}{$\uparrow$}\\
CC~\cite{peng2019correlation}  &72.97$\pm$0.17 &71.96$\pm$0.11\textcolor{red}{$\downarrow$} &72.21$\pm$0.25&70.64$\pm$0.12\textcolor{red}{$\downarrow$} &71.14$\pm$0.06&70.85$\pm$0.12\textcolor{red}{$\downarrow$}\\
VID~\cite{ahn2019variational}  &73.09$\pm$0.21 &73.48$\pm$0.23\textcolor{green}{$\uparrow$} &73.30$\pm$0.13 &73.06$\pm$0.17\textcolor{green}{$\downarrow$}&73.38$\pm$0.09&75.80$\pm$0.12\textcolor{green}{$\uparrow$}\\
RKD~\cite{park2019relational}  &71.90$\pm$0.11 &72.50$\pm$0.23\textcolor{green}{$\uparrow$} &72.22$\pm$0.20&72.99$\pm$0.17\textcolor{green}{$\uparrow$} &72.28$\pm$0.39&73.38$\pm$0.09\textcolor{green}{$\uparrow$}\\
PKT~\cite{passalis2018learning}&73.64$\pm$0.18 &75.24$\pm$0.15\textcolor{green}{$\uparrow$} &73.45$\pm$0.19&74.29$\pm$0.18\textcolor{green}{$\uparrow$} &74.10$\pm$0.25&76.50$\pm$0.12\textcolor{green}{$\uparrow$}\\
AB~\cite{heo2019knowledge}     &73.17$\pm$0.31&72.34$\pm$0.09\textcolor{red}{$\downarrow$}  &72.38$\pm$0.31&72.88$\pm$0.18\textcolor{green}{$\uparrow$} &73.55$\pm$0.31&73.33$\pm$0.12\textcolor{red}{$\downarrow$}\\
FT~\cite{kim2018paraphrasing}  &72.86$\pm$0.12&71.57$\pm$0.11\textcolor{red}{$\downarrow$}  &71.59$\pm$0.15&71.47$\pm$0.23\textcolor{red}{$\downarrow$} &71.75$\pm$0.20&72.81$\pm$0.21\textcolor{green}{$\uparrow$}\\
NSP~\cite{huang2017like}       &73.30$\pm$0.28 &72.06$\pm$0.20\textcolor{red}{$\downarrow$} &72.24$\pm$0.22&72.11$\pm$0.15\textcolor{red}{$\downarrow$} &74.12$\pm$0.19&N/A\\
CRD~\cite{tian2019contrastive} &75.51$\pm$0.18 &73.84$\pm$0.14\textcolor{red}{$\downarrow$} &74.14$\pm$0.22&73.54$\pm$0.19\textcolor{red}{$\downarrow$} &75.11$\pm$0.32&76.70$\pm$0.16\textcolor{green}{$\uparrow$}\\
\hline
\bf \shortname &\textbf{75.92}$\pm$0.19&\textbf{76.24}$\pm$0.02\textcolor{green}{$\uparrow$} &\textbf{74.75}$\pm$0.20&\textbf{75.32}$\pm$0.06\textcolor{green}{$\uparrow$} &\textbf{76.40}$\pm$0.13&\textbf{77.40}$\pm$0.17\textcolor{green}{$\uparrow$}\\
\hline
Teacher &79.42 &- &75.61&- &79.42&-\\
\hline
\end{tabular}
\label{tab:un_kds}
\end{table*}

\vspace{0.1cm}
\noindent {\bf Results: }
We report the results in Table~\ref{tab:un_kds}. 
We make the following observations: 
{\bf (1)} Whilst KD~\cite{hinton2015distilling}, RKD~\cite{park2019relational}, PKD~\cite{passalis2018learning}, and our {\shortname} consistently improve from using unlabeled training data, FitNet~\cite{romero2015fitnets} and CC~\cite{peng2019correlation} degrade across all different networks.
This implies that intermediate feature matching~\cite{romero2015fitnets} and
inter-instance correlation~\cite{peng2019correlation} are less robust to unconstrained data.
This further verifies that using the high-level feature alignment as in \shortname{} is more reliable.
{\bf (2)} Besides, AT~\cite{zagoruyko2016paying}, SP~\cite{tung2019similarity}, AB~\cite{heo2019knowledge}, FT~\cite{kim2018paraphrasing}, VID~\cite{ahn2019variational}, and CRD~\cite{tian2019contrastive} not necessarily improve, conditioned on network pairs.
In particular, heterogeneous architectures are preferred by CRD and FT, whilst others present no clear trend.
{\bf (3)} Our {\shortname} yields consistently best results, with clear boost up from open-set unlabeled data across all the cases. This suggests the overall performance advantage and network robustness of our model design despite its simplicity.
{\bf (4)} Interestingly, the logit-matching KD \cite{hinton2015distilling}
benefits more from unlabeled data than the other competitors.
However, its overall performance is inferior to ours.
In a nutshell, this test shows that not all distillation methods can easily benefit from using open-set unlabeled data, with some methods even conditioned on the network choice.

\vspace{0.1cm}
\noindent {\bf Integration with self-supervised learning: }
Self-supervised learning is an orthogonal dimension for improving knowledge distillation. 
For instance, a recent method SSKD \cite{SSKD}. integrates self-supervised learning (\eg, instance discrimination by contrastive learning \cite{chen2020simple}) to distillation,
conceptually complementary to our \shortname{}.
We evaluate this combination in our open-set SSL setting with Tiny-ImageNet as the unlabeled data.
From Table \ref{tab:sskd}, we observe that both SSKD and our \shortname{} in isolation are effective, and importantly, they can be integrated well for further improvement. 

\begin{table}[h]
\centering
\caption{Integration with self-supervised learning. Labeled set: CIFAR-100; Unlabeled set: Tiny-ImageNet. Metric: Top-1 accuracy (\%).}
\resizebox{1\linewidth}{!}{
\setlength\tabcolsep{3.5pt}
\begin{tabular}{l|c|c|c}
\hline
Network& ResNet8$\times$4 &  WRN40-1 & ShuffleNetV1\\
\hline\hline
\em Supervised learning &72.50 &71.98 &70.50\\ \hline
SSKD~\cite{SSKD} &75.15&74.64&78.36\\
\bf{SRD}  &76.24&75.32&77.40\\
SSKD+\bf{SRD} &\bf{76.38}&\bf{75.61}&\bf{78.74}\\\hline
Teacher&79.42&75.61&79.42\\
\hline
\end{tabular}}
\label{tab:sskd}
\vspace{-4mm}
\end{table}


\subsection{Evaluation on Semi-Supervised Learning Methods}\label{sec:res_ukds}

\noindent {\bf Setting: }
In addition to distillation methods, we further compare our \shortname{} with existing SSL methods capable of leveraging unlabeled data.
To enable comparing the two different approaches, we adopt the same training and test setting as in Sec.~\ref{ssec:CIFAR-100}.
In particular, we adopt the same three student networks as the target models: ResNet8$\times$4, WRN40-1 and ShuffleNetV1.
For \shortname{}, we apply the same three (teacher, student) pairs (see top of Table~\ref{tab:un_kds}).
Following the convention of SSL, we report the average result of the last 20 epochs.
\revision{Specifically, our experimental design utilizes CIFAR-100 as the source of labeled data, while the unlabeled datasets include classes not found in CIFAR-100, sourced from TinyImageNet \cite{le2015tiny}, Places365 \cite{places}, and CC3M \cite{sharma2018conceptual}. The selection of unlabeled data from three distinct sources introduces a significant variability in relevance and complexity: TinyImageNet shares some similarities with CIFAR-100, making it somewhat related; Places365, with its focus on 365 different scene categories, presents a stark contrast to the object-centric images of TinyImageNet and CIFAR-100; CC3M, an even less refined dataset, is amassed without stringent filtering or manual labeling. To categorize the unlabeled datasets, we utilized a pretrained ResNet32 × 4 classifier to label each sample. Any sample with a confidence score below 0.9 was deemed OOD. Through this process, we determined that 73\% of Tiny-ImageNet (from a total of 100,000 samples), 70\% of Places365 (from a total of 1,803,460 samples), and 78\% of CC3M (from a total of 2,313,472 samples) fall into the OOD category. When incorporating these datasets as unlabeled data alongside CIFAR-100, the proportions of OOD data vary, ranging from 48\% for TinyImageNet and 68\% for Places365 to 76\% for CC3M.}


\vspace{0.1cm}
\noindent {\bf Competitors: }
We evaluate four representative closed-set
(PseudoLabel~\cite{lee2013pseudo},
MeanTeacher~\cite{tarvainen2017mean},
MixMatch~\cite{berthelot2019mixmatch} and 
FixMatch~\cite{sohn2020fixmatch})
and three state-of-the-art open-set 
(MTCR \cite{yu2020multi}, T2T~\cite{T2T} and OpenMatch~\cite{openmatch})
SSL methods. 
%
For each method, we adopt its well-tuned default setting provided in the respective source codes. 
We also include the {\em supervised learning} baseline
without using any unlabeled data.

\begin{table}[h]
\caption{Comparing state-of-the-art closed-set and open-set semi-supervised learning methods.
{\em Labeled set}: CIFAR-100;
{\em Unlabeled set}: Tiny-ImageNet.
{\em Metric}: Best Top-1 accuracy (\%). 
}
\centering
\resizebox{1\linewidth}{!}{
\setlength\tabcolsep{3.5pt}
\begin{tabular}{l|c|c|c}
\hline
Network   & ResNet8$\times$4 &WRN40-1 &ShuffleNetV1\\
\hline \hline
\em Supervised learning &72.50 &71.98 &70.50\\
\hline
PseudoLabel~\cite{lee2013pseudo}&33.33&54.48&42.47\\
MeanTeacher~\cite{tarvainen2017mean}&70.18&70.25&65.60\\
FixMatch~\cite{sohn2020fixmatch}&68.85&65.54&61.27\\
MixMatch~\cite{berthelot2019mixmatch} &\underline{74.27} &\underline{74.01} &\underline{75.79}\\
\hline
MTCR~\cite{yu2020multi}&65.42&61.84&42.34\\
T2T~\cite{T2T}         &66.05&62.55 &57.53\\
OpenMatch~\cite{openmatch}&70.41&69.08&66.88\\
\hline
{\bf \shortname}  &\bf{76.24}&\bf{75.32}&\bf{77.40}\\
\hline
\end{tabular}}
\label{tab:ssl_tin}
\vspace{-4mm}
\end{table}

\vspace{0.1cm}
\noindent {\bf Results: }
The results are compared in Table~\ref{tab:ssl_tin}.
We draw several interesting observations.
{\bf (1)} With a strong closed-set label space assumption, as expected pseudo labeling \cite{lee2013pseudo} gives the poorest performance with significant degradation from the supervised baseline over all three networks. 
This is because assigning a label for every unlabeled sample in unconstrained open-set setting is highly error-prone.
In practice, it is found that $\sim90\%$ of unlabeled data were wrongly labeled to a single class ``{\em crocodile}'' during training and $\sim50\%$ of test samples were classified as ``{\em crocodile}'' mistakenly.
{\bf (2)} 
Relying on less rigid consistency regularization,
MeanTeacher \cite{tarvainen2017mean} yields better performance but still suffers from accuracy drop compared to using only labeled training data.
This reveals the limitation of conventional consistency loss in tackling open-set unlabeled data.
{\bf (3)} Combining consistency regularization and pseudo-labeling, the recent closed-set SSL method FixMatch \cite{sohn2020fixmatch} reasonably performs at a level between \cite{tarvainen2017mean} and \cite{lee2013pseudo}. 
Note, this is rather different from its original closed-set SSL results, revealing previously unseen limitation of this hybrid strategy.
{\bf (4)} 
Among all the closed-set SSL methods,
MixMatch \cite{berthelot2019mixmatch} is the only one capable of achieving accuracy gain from unconstrained unlabeled data.
This indicates a new finding that multi-augmentation pooling and sharpening based consistency turns out to be more generalizable to unconstrained open-set SSL.
{\bf (5)} 
Surprisingly, it is observed that
all open-set SSL methods (MTCR \cite{yu2020multi}, T2T \cite{T2T} and OpenMatch \cite{openmatch}) fail to improve over the supervised learning baseline.
This contradicts the reported findings under their simpler and less realistic settings including fewer target classes and unlabeled samples but higher class overlap percentages between labeled and unlabeled sets.
This presents a failure case for previous open-set SSL methods and would be thought-provoking and inspire more extensive investigation under more challenging and realistic scenarios with abundant unlabeled data.
{\bf (6)} 
Our \shortname{} consistently improves the performance of all three networks and surpasses significantly the best competitor MixMatch \cite{berthelot2019mixmatch}. 
This validates a clear advantage of the proposed semantic representational distillation over previous SSL methods in handling open-set unlabeled data.
Conceptually, \shortname{} can be also viewed as {\em semantic consistency regularization} derived from a pre-trained teacher.
Our design shares the general consistency regularization spirit whilst differentiating from typical data augmentation based consistency in formulation \cite{berthelot2019mixmatch,sohn2020fixmatch}.
{\bf (7)} 
Regarding the high-level approach,
distillation methods (Table \ref{tab:un_kds}) are shown to be generally superior over both closed-set and open-set SSL ones (Table \ref{tab:ssl_tin}). This implies that using a pretrained teacher model as semantic guidance in various ways could be more advantageous than OOD detection of open-set SSL methods in capitalizing unconstrained unlabeled data.
Actually, it is shown that detecting and discarding OOD samples even results in harmful effect over the ignorance of OOD data typical with closed-set SSL methods.

\noindent
\revision{{\bf Integration with semi-supervised learning: }
Knowledge Distillation (KD) and open-set Semi-Supervised Learning (SSL) represent two different paradigms in machine learning,
with distinct pipeline designs and research objectives. We concur that integrated evaluation may cast additional insights.
Specifically, we have now conducted an experiment by integrating our SRD (the knowledge distillation component) with the best performing SSL method MixMatch.
MixMatch operates by generating low-entropy, or high-confidence, labels for augmented versions of unlabeled data. Subsequently, we combine both labeled and unlabeled data by the application of MixUp. 
We have extended this process by applying SRD to both labeled and unlabeled data, under the strategic guidance of a teacher network. 
As shown in Table \ref{tab:mixmatch}, we observe that both MixMatch and SRD in isolation are effective, and importantly, their combination achieves further improvement.}
\begin{table}[h]
\centering
\caption{Integration with semi-suppervised learning. Labeled set: CIFAR-100; Unlabeled set: Tiny-ImageNet. Metric: Top-1 accuracy (\%).}
\resizebox{1\linewidth}{!}{
\setlength\tabcolsep{3.5pt}
\begin{tabular}{l|c|c|c}
\hline
Network& ResNet8$\times$4 &  WRN40-1 & ShuffleNetV1\\
\hline\hline
\em Supervised learning &72.50 &71.98 &70.50\\ \hline
\bf{SRD} &76.24&75.32 &77.40\\
MixMatch  &74.27 &74.01 &75.79\\
MixMatch+\bf{SRD} &\bf{77.13}&\bf{76.06} &\bf{78.25}\\\hline
Teacher&79.42&75.61&79.42\\
\hline
\end{tabular}
\label{tab:mixmatch}
}
\vspace{-4mm}
\end{table}

\noindent {\bf How does OOD detection fail? }
To further examine the failure of open-set SSL methods, we analyse the behavior of OOD detection with T2T \cite{T2T} and
OpenMatch \cite{openmatch} during training.
In particular, we track the per-epoch proportion of unlabeled data passing through the OOD detector, with their ground-truth class labels categorized into OOD and in-distribution (IND).
It is shown in Fig. \ref{fig:t2t_tin} that both T2T and
OpenMatch can identify the majority of OOD samples at most time whilst keeping away from IND samples at varying rates.
However, their performance is still inferior to supervised learning baseline. 
To further isolate the performance factors,
we test particularly the supervised learning component of OpenMatch by deactivating all the unsupervised loss terms. 
We find that its use of unlabeled data turns out to degrade the performance. 
This is due to both the challenge of identifying OOD samples under such more difficult open-set settings as studied here,
and improper use of unlabeled data.
We conjugate that in highly unconstrained open-set SSL scenarios, the OOD strategy has become ineffective.
On the contrary, with KD a pretrained teacher can instead extract more useful latent knowledge (\eg, parts and attributes shared across labeled and unlabeled classes) from unlabeled data not limited to labeled/known classes and more effectively improve the model optimization.

\begin{figure}[ht]
\centering
\subfigure[\fontsize{6}{7}\selectfont{T2T w/ ResNet8$\times$4}]{\includegraphics[width=0.32\linewidth]{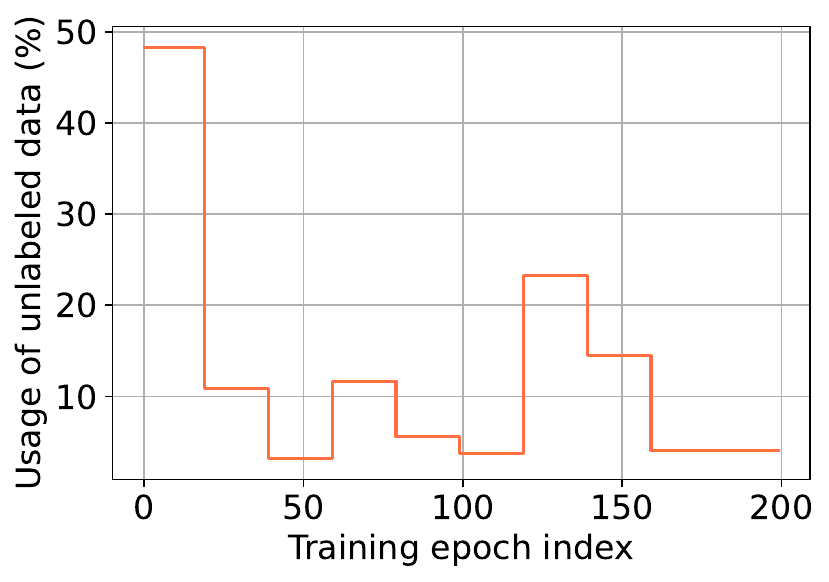}}
\subfigure[\fontsize{6}{7}\selectfont{T2T w/ WRN40-1}]{\includegraphics[width=0.32\linewidth]{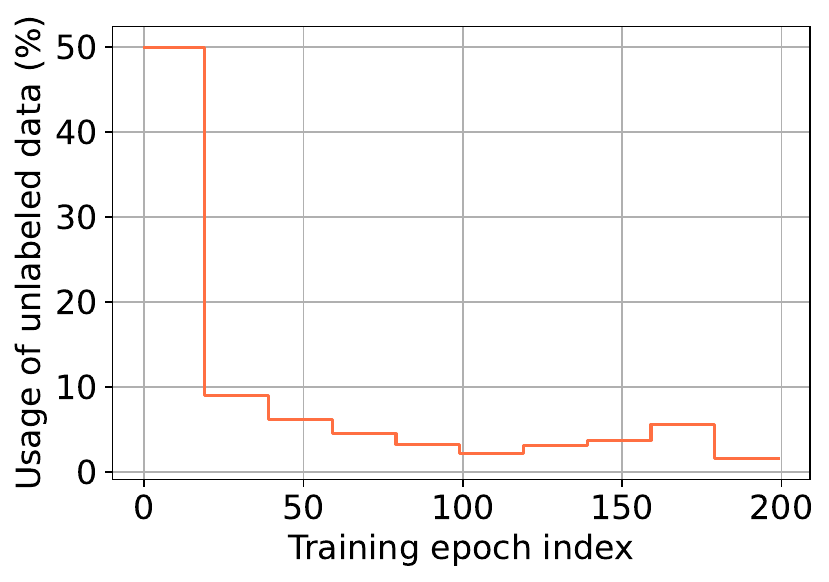}} 
\subfigure[\fontsize{6}{7}\selectfont{T2T w/ ShuffleNetV1}]{\includegraphics[width=0.32\linewidth]{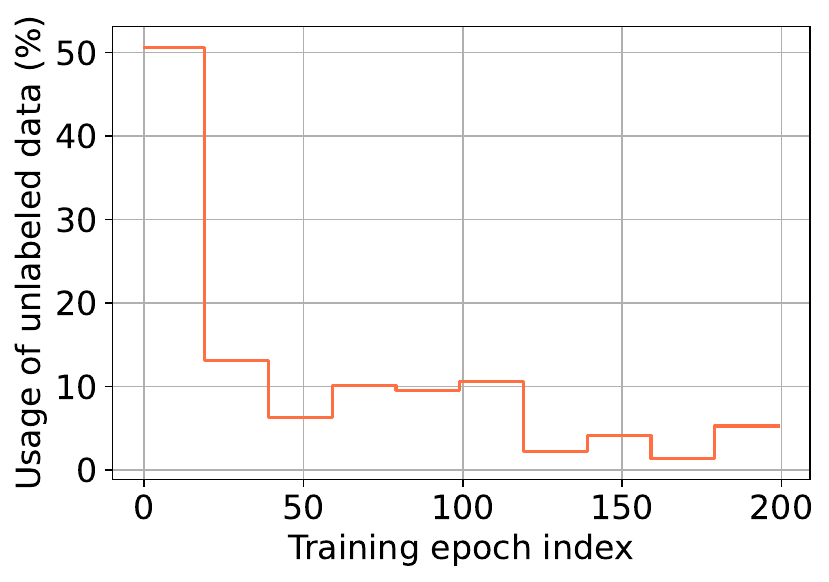}}\\
\subfigure[\fontsize{6}{7}\selectfont{OM w/ ResNet8$\times$4}]{\includegraphics[width=0.32\linewidth]{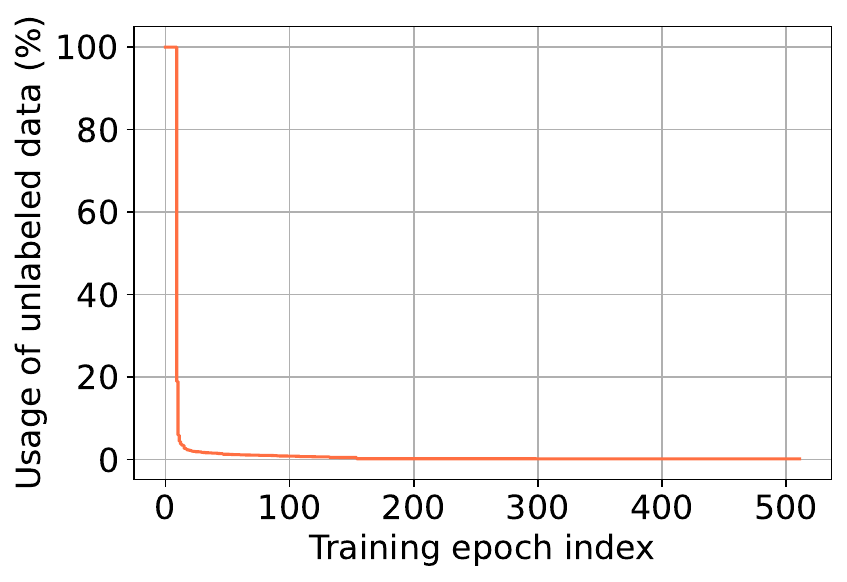}}
\subfigure[\fontsize{6}{7}\selectfont{OM w/ WRN40-1}]{\includegraphics[width=0.32\linewidth]{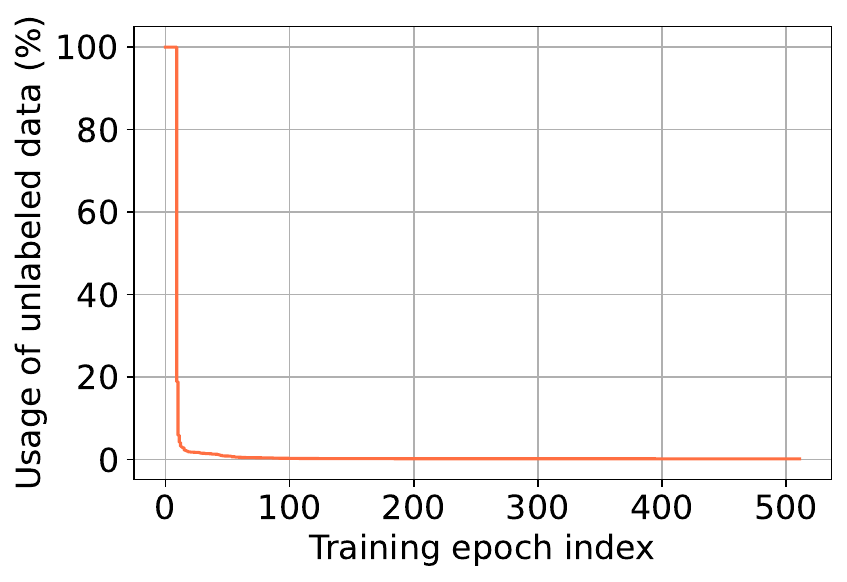}} 
\subfigure[\fontsize{6}{7}\selectfont{OM w/ ShuffleNetV1}]{\includegraphics[width=0.32\linewidth]{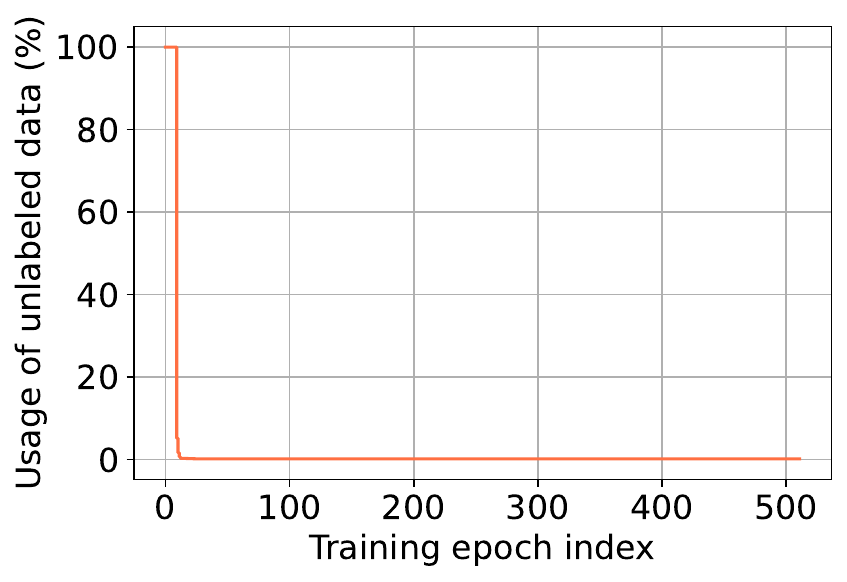}}
\caption{Per-epoch usage of unlabeled data (those surviving through OOD detection) with ({\em top}) T2T \cite{T2T} and ({\em bottom}) OpenMatch (OM) \cite{openmatch}. 
}
\label{fig:t2t_tin}
\vspace{-4mm}
\end{figure}

\subsection{Further Analysis}


We further conduct a series of 
analytical experiments for providing in-depth insights in model design, performance evaluation, training data, image resolution under open-set SSL setting.

\vspace{0.1cm}
\noindent {\bf Effect of unlabeled data size:}
We evaluate the effect of unlabeled data size on the performance.
From the training set of Tiny-ImageNet, we create four varying-size (\{25\%, 50\%, 75\%, 100\%\}) unlabeled sets via random and teacher prediction score based selection, respectively.
As shown in Fig.~\ref{fig:amount}, both KD and \shortname{} benefit from more unlabeled data regardless of the selection process.
This suggests they are generally scalable and insensitive to unlabeled data filtering.

\begin{figure}[h]
\centering
\includegraphics[width=0.99\linewidth]{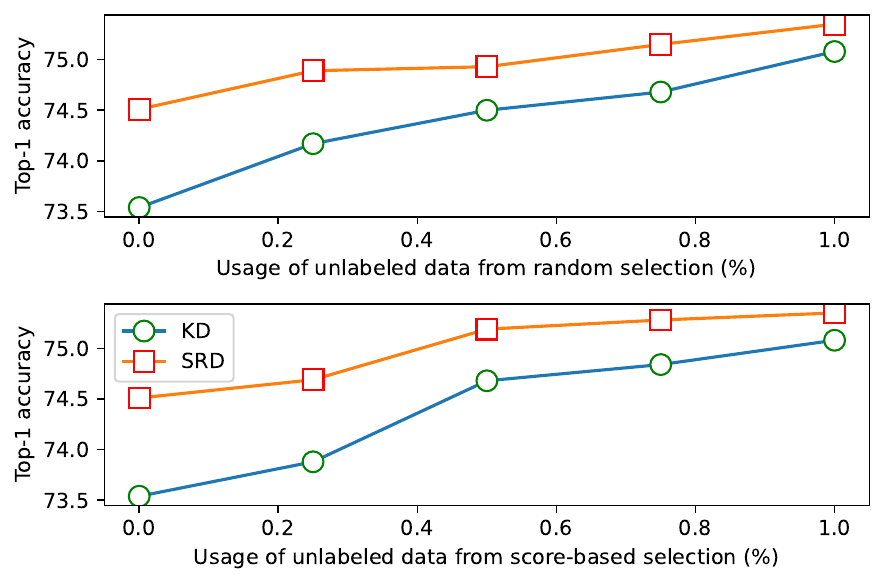}
\caption{
Size effect of unlabeled data selected ({\em Top}) randomly 
or ({\em Bottom}) by the teacher prediction score.
Teacher: WRN40-2.
Student: WRN40-1.}
\label{fig:amount}
\end{figure}

\vspace{0.1cm}
\noindent
\added{\bf{Effect of labeled data size: }}
\added{We evaluate the impact of the amount of labeled data on performance.
For our experiment, we randomly select each class 250 images from the training set of CIFAR-100 for training. Evaluation remains unchanged.
It is shown in Table \ref{tab:labeled_size} that the deduction of in the quantity of labeled data results in a decrease of performance.
Despite this, {\shortname{}} still ranks first among the competitors.
This indicates that the effectiveness of the model generally scales with the size of the labeled data.}
\begin{table}[h]
\centering
\added{
\caption{Effect of labeled training set size. Labeled set: 50\% CIFAR-100; Unlabeled set: Tiny-ImageNet. Metric: Top-1 accuracy (\%).}
\begin{tabular}{l|c}
\hline
Network&  WRN40-1 \\
\hline
\em supervised learning &65.65\\
\hline
MixMatch~\cite{berthelot2019mixmatch} &68.71\\
OpenMatch~\cite{openmatch} &62.80\\
\hline
KD~\cite{hinton2015distilling}&69.65\\ 
\bf {\shortname{}} &\bf{69.97}\\
\hline
Teacher &69.41\\
\hline
\end{tabular}
\label{tab:labeled_size}}
\vspace{-4mm}
\end{table}

\vspace{0.1cm}
\noindent {\bf OOD detection: }
It might be interesting to see how OOD works with distillation methods. 
To that end, we adopt the OOD detector of T2T \cite{T2T} which is a binary classifier with the teacher's representation, trained jointly during distillation. 
It is shown in Table \ref{tab:KD_OOD} that
OOD detection has very marginal effect in most cases, although more unlabeled data on average (Fig. \ref{fig:kds_t2t_tin}) are selected as compared to its original form (Fig. \ref{fig:t2t_tin}).
This indicates little complementary between OOD detection and distillation methods.

\begin{figure}[ht]
\centering
\subfigure[\fontsize{5}{7}\selectfont{KD+OOD w/ ResNet8$\times$4}]{\includegraphics[width=0.32\linewidth]{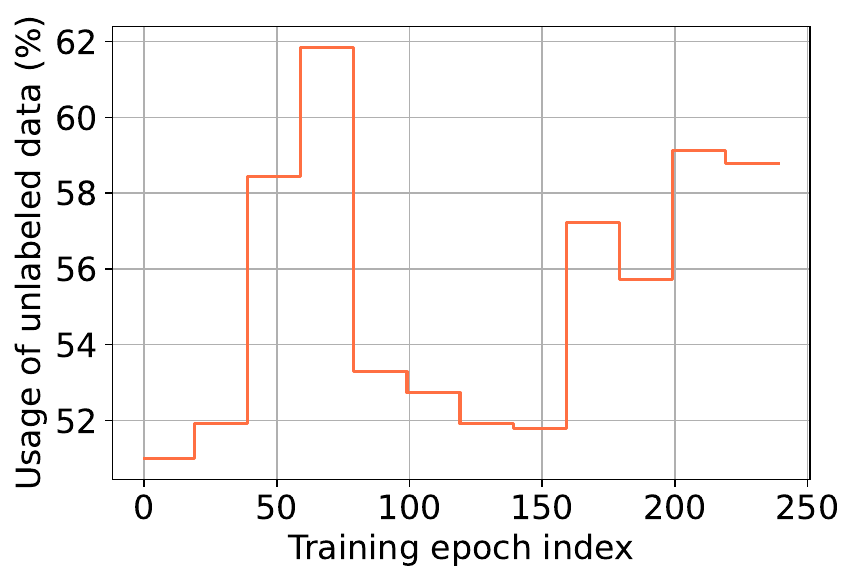}}
\subfigure[\fontsize{5}{7}\selectfont{KD+OOD w/ WRN40-1}]{\includegraphics[width=0.32\linewidth]{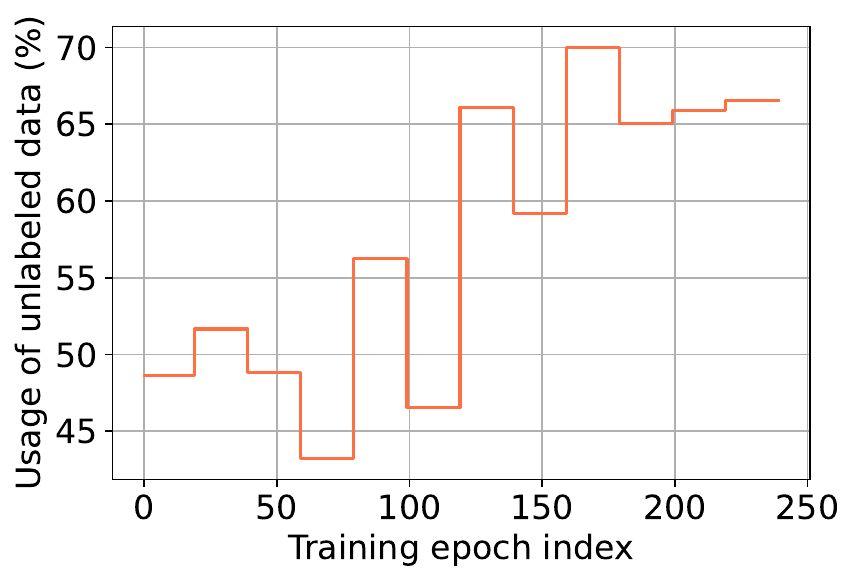}} 
\subfigure[\fontsize{5}{7}\selectfont{KD+OOD w/ ShuffleNetV1}]{\includegraphics[width=0.32\linewidth]{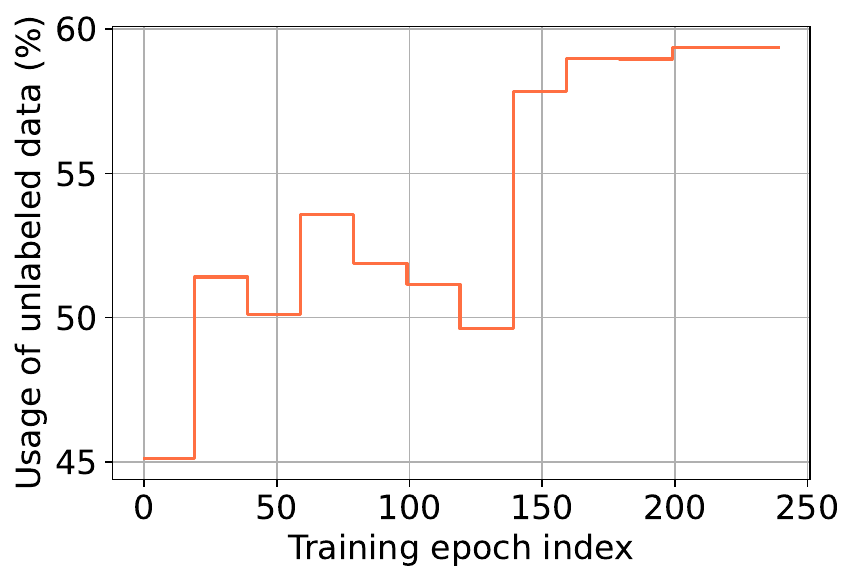}}\\
\subfigure[\fontsize{5}{7}\selectfont{SRD+OOD w/} ResNet8$\times$4]{\includegraphics[width=0.32\linewidth]{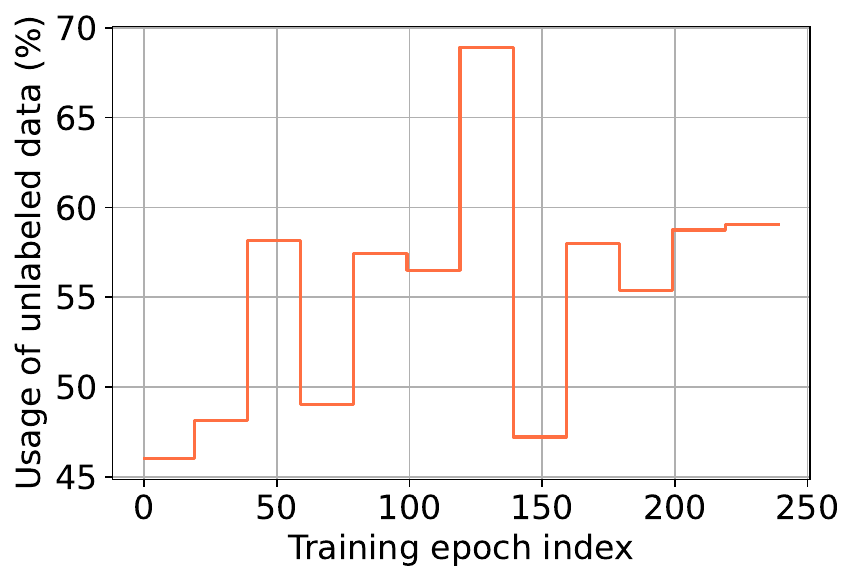}}
\subfigure[\fontsize{5}{7}\selectfont{SRD+OOD w/ WRN40-1}]{\includegraphics[width=0.32\linewidth]{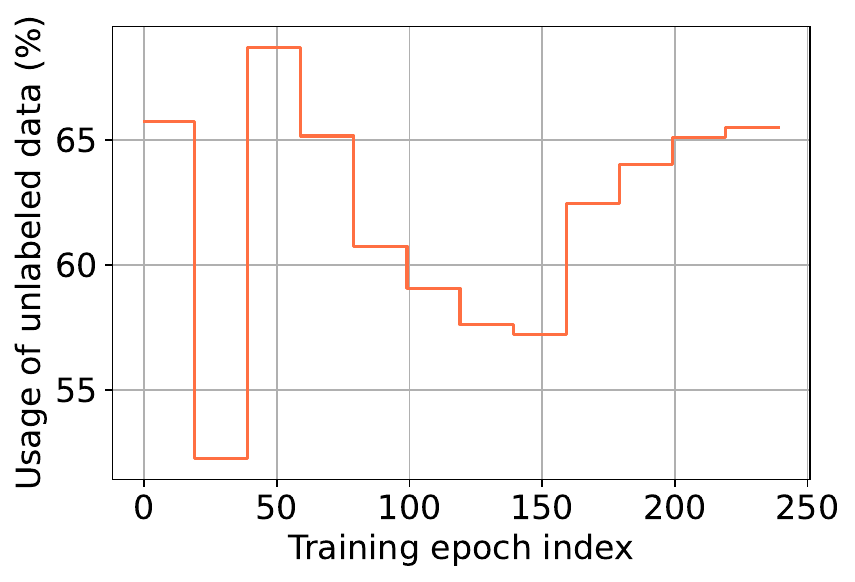}} 
\subfigure[\fontsize{5}{7}\selectfont{SRD+OOD w/ ShuffleNetV1}]{\includegraphics[width=0.32\linewidth]{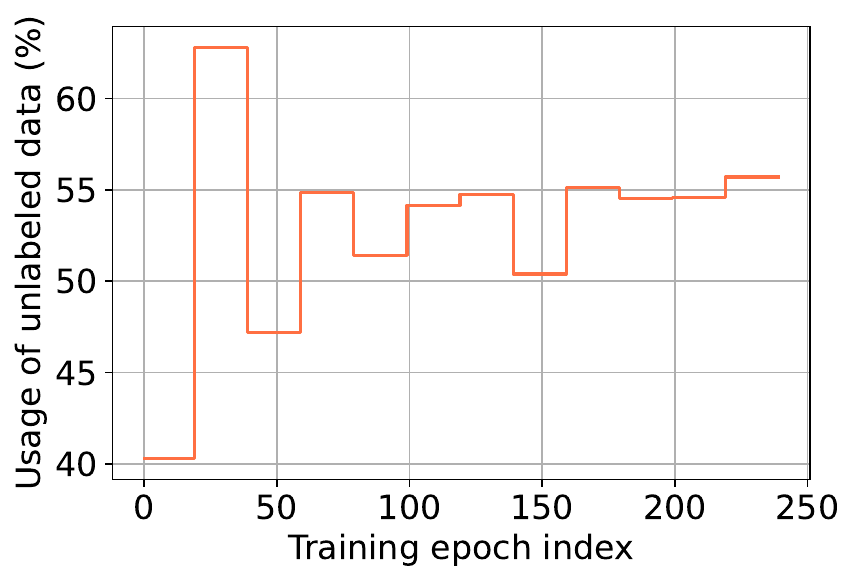}}
\caption{The per-epoch usage of unlabeled data 
when an OOD detector is equipped with ({\em top}) KD \cite{hinton2015distilling} and ({\em bottom}) our \shortname{}. }
\label{fig:kds_t2t_tin}
\vspace{-4mm}
\end{figure}


\begin{table}[h]
\caption{Effect of OOD detection on knowledge distillation methods.}
\centering
\begin{tabular}{l|c|c|c}
\hline
Network& ResNet$8\times4$ &  WRN40-1 & ShuffleNetV1\\
\hline
KD~\cite{hinton2015distilling} &\bf 74.68 &75.08 &76.53\\ 
KD+OOD      &74.58  &\bf 75.31  &\bf 76.54 \\ \hline
\bf \shortname  &\textbf{76.24} &75.32 &77.40\\
{\bf \shortname}+OOD  &76.08  &\bf{75.54}  &\textbf{77.41}\\
\hline
\end{tabular}
\label{tab:KD_OOD}
\vspace{-4mm}
\end{table}

\vspace{0.1cm}
\noindent
\textbf{Data augmentation based consistency: }
A key component with SSL methods is data augmentation based consistency regularization.
Here we examine how it can work together with distillation loss which instead imposes cross-network consistency.
We experiment with stochastic augmentation operations including random cropping by up to 4-pixels shifting and random horizon flipping.
Given an unlabeled image, we generate two views via data augmentation
and feed one view into the teacher and both into the student.
A consistency loss is then applied via maximizing their logit similarity.
It is shown in Table \ref{tab:KD_CS} that data augmentation based consistency would lead to performance degradation.
This suggests that cross-network and cross-augmentation consistency 
are incompatible with each other.
As we find, this is because the teacher could output rather different predictions for the two views of a single image (\eg, giving different predicted labels on $\sim$40\% of unlabeled images), causing contradictory supervision signal.

\begin{table}[h]
\caption{Effect of data augmentation based consistency (DAC) on knowledge distillation methods.}
\centering
\begin{tabular}{l|c|c|c}
\hline
Network& ResNet8$\times$4 &  WRN40-1 & ShuffleNetV1\\
\hline
KD~\cite{hinton2015distilling}     &\bf 74.68 &\bf 75.08  &\bf 76.52\\
KD~\cite{hinton2015distilling}+DAC  &72.82 &74.34  &75.64
\\ \hline
\bf \shortname                         &\bf 76.24 &\bf 75.32 &\bf 77.40\\
{\bf \shortname}+DAC                      &75.66 &73.70  &76.34\\
\hline
\end{tabular}
\label{tab:KD_CS}
\vspace{-4mm}
\end{table}

\vspace{0.1cm}
\noindent
\textbf{More unconstrained unlabeled data: }
For evaluating the generality and scalability in terms of unlabeled data, we test less related unlabeled data 
by replacing Tiny-ImageNet with {\em Places365} \cite{places}.
This dataset has 1,803,460 training images from 365 scene categories, drastically different from object-centric images from Tiny-ImageNet and CIFAR-100 (see Fig. \ref{fig:plate}). 
This hence presents a more challenging open-set SSL scenario.
Similarly, we use its training set as unlabeled data.  
All the other settings remain.
We make similar observations from Table \ref{tab:KD_place}.
{\bf (1)}
Similarly, all closed-set SSL methods except MixMatch fail to improve over supervised learning baseline.
{\bf (2)}
Again, open-set SSL methods are all ineffective and even suffer from more performance drop.
{\bf (3)}
Our \shortname{} consistently delivers the best accuracy with a decent margin over the conventional distillation method.
{\bf (4)}
Overall, distillation methods remain superior over all SSL competitors, suggesting their generic advantages even at more challenging scenarios with less relevant unlabeled data involved.

\begin{figure}[h]
    \centering
    \includegraphics[width=0.9\linewidth]{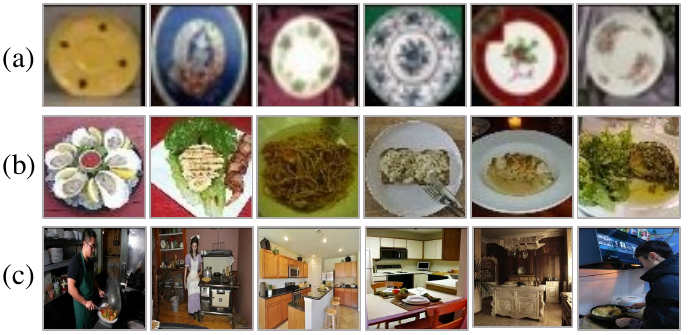}
    \caption{Object-centric images from ({\bf a}) {CIFAR-100} and  ({\bf b}) {Tiny-ImageNet} \vs{} scene images from ({\bf c}) {Place365} including plate objects.}
    \label{fig:plate}
    \vspace{-4mm}

\end{figure}

\begin{table}[h]
\centering
\caption{Generality and scalability test with unconstrained unlabeled data from the Place365 dataset.}
\resizebox{1\linewidth}{!}{
\setlength\tabcolsep{3.5pt}
\begin{tabular}{l|c|c|c}
\hline
Network& ResNet8$\times$4 &  WRN40-1 & ShuffleNetV1\\
\hline
\em Supervised learning &72.50 &71.98 &70.50\\ \hline
PseudoLabel~\cite{lee2013pseudo}&21.62&49.96&40.43\\
MeanTeacher~\cite{tarvainen2017mean}&66.94&53.19&60.32\\
FixMatch~\cite{sohn2020fixmatch}&68.57&66.42&64.26\\
MixMatch~\cite{berthelot2019mixmatch}&73.40&73.72&74.89\\
\hline
MTCR~\cite{yu2020multi}&58.08&55.38&33.67\\
T2T~\cite{T2T}&63.22&61.88&61.54\\
OpenMatch~\cite{openmatch}&58.55&69.80&67.57\\
\hline
KD~\cite{hinton2015distilling}    &74.13   & 74.29 &75.81\\
\bf {\shortname}  &\bf{75.93}   & \bf{75.40} &\bf{77.18}\\

\hline
\end{tabular}}
\label{tab:KD_place}
\vspace{-4mm}
\end{table}

Considering that TinyImageNet and Place365 both were created under filtering and manual annotation, we test a further less curated dataset, CC3M \cite{sharma2018conceptual}, as unlabeled data. 
We compare \shortname{} with the top semi-supervised and distillation competitors: MixMatch \cite{berthelot2019mixmatch}, OpenMatch \cite{openmatch}, and KD \cite{hinton2015distilling}.
As shown in Table \ref{tab:KD_cc3m}, we obtain consistent results 
as on Tiny-ImageNet and Place365. 
This implies that image curation has little impact in such open-set scenarios with good scales. 
This is not surprising since manual annotations are not used and image selection would not significantly reduce the open-set challenges.
\begin{table}[h]
\centering
\caption{Generality and scalability test with more unconstrained unlabeled data from the CC3M dataset.}
\resizebox{1\linewidth}{!}{
\setlength\tabcolsep{3.5pt}
\begin{tabular}{l|c|c|c}
\hline
Network& ResNet8$\times$4 &  WRN40-1 & ShuffleNetV1\\
\hline\hline
\em Supervised learning &72.50 &71.98 &70.50\\ \hline
MixMatch~\cite{berthelot2019mixmatch}&73.01&73.02&74.16\\ 
OpenMatch~\cite{openmatch}&71.56&69.61&66.60\\
KD~\cite{hinton2015distilling} &73.78   &74.20 &75.22\\\hline
\bf {\shortname}  &\bf{76.24}   &\bf{74.99} &\bf{76.65}\\
\hline
\end{tabular}}
\label{tab:KD_cc3m}
\end{table}

\vspace{0.1cm}
\noindent{\bf More performance metrics:}
\added{We further adopt multi-class Area Under the Receiver Operating Characteristic (AUROC) for evaluation. In our cases, we employ the standard scikit toolbox\footnote{ \url{https://scikit-learn.org/stable/auto_examples/model_selection/plot_roc.html}} which treats the multi-class classification task as a set of binary classification task across each class. 
We compare \shortname{} with MixMatch \cite{berthelot2019mixmatch}, OpenMatch \cite{openmatch}, and KD \cite{hinton2015distilling}.
As show in Figure~\ref{fig:auroc}, SRD is consistently superior to the competitors.}
\begin{figure}[h]
    \centering
    \includegraphics[width=0.9\linewidth]{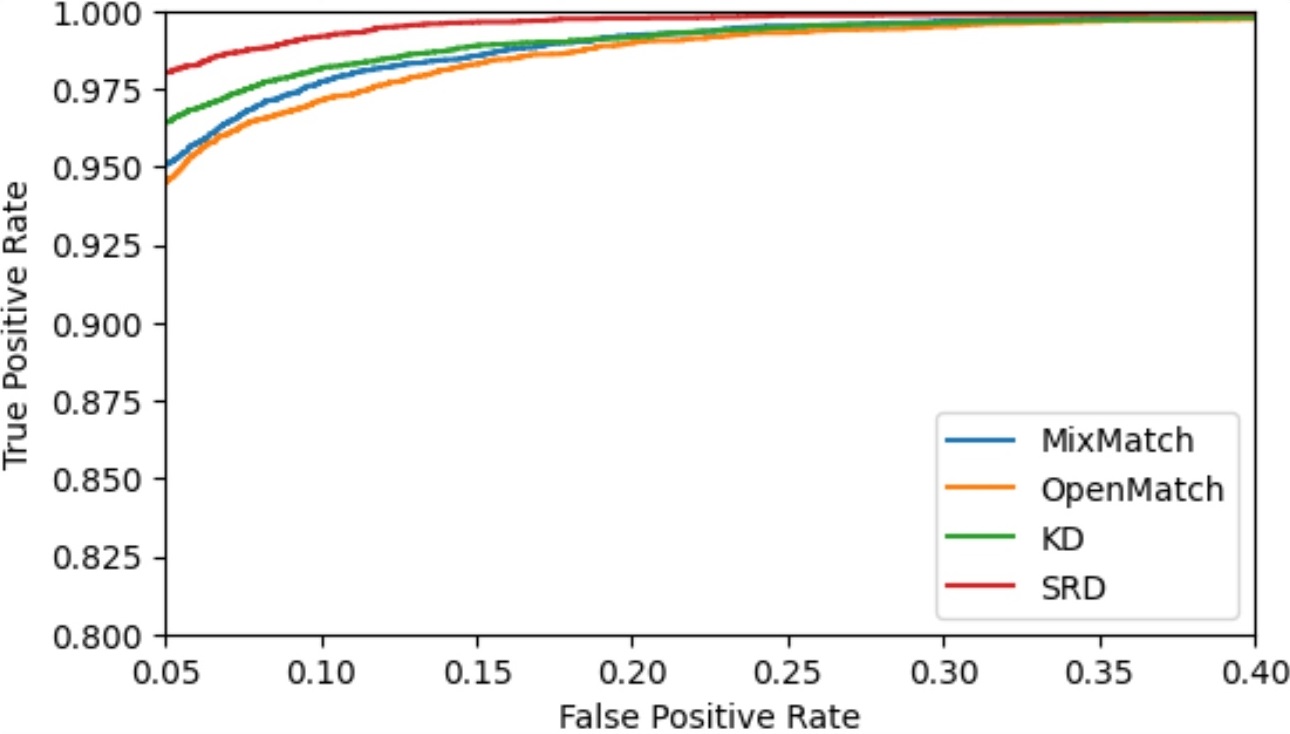}
    \caption{\added{Area under the receiver operating characteristic (AUROC) curve on CIFAR-100 test set.}}
    \label{fig:auroc}
\end{figure}

\vspace{0.1cm}
\noindent
\textbf{\revision{Effect of higher image resolution: }}
\label{sec:imagenet_unlabeled}
\revision{
To assess our method's effectiveness with high-resolution images, we test a subset of ImageNet-1K (ImageNet-Sub), which comprises 500 classes with 500 samples per class. Evaluations are conducted on the original ImageNet evaluation dataset, focusing on the same 500 classes. Additionally, for the unlabeled data, we select a subset (YFCC-Sub) from the Yahoo Flickr Creative Commons 100 Million (YFCC-100M) \cite{thomee2016yfcc100m} dataset, which is four times the size of the labeled data with OOD ratio of 80\%. The models are trained with SGD as optimizer, weight decay $1e-3$ and a standard learning rate of 0.1, and decayed by 0.1 every 30 epochs for total 100 epochs. The batch size is set to 256. Specifically, we employ a ResNet50 model as the teacher, guiding two student models with distinct architectures: ResNet18 and MobileNet.
}

\revision{As shown in Table \ref{tab:imagenetsub}, we obtain consistent results as on Tiny-ImageNet, Place365 and CC3M. This implies that image resolution will not affect the overall conclusion.}

\begin{table}[h]
\centering
\revision{
\caption{Generality on higher resolution images. {\em Labeled set}: ImageNet-sub, {\em Unlabeled set}:: YFCC-Sub. Metric: Top-1 \%. Teacher: ResNet50.}
\begin{tabular}{l|c|c}
\hline
Network& ResNet18 & MobileNet\\
\hline\hline
\em Supervised learning &63.49 &65.10\\ \hline
MixMatch~\cite{berthelot2019mixmatch}&65.46&66.31\\
OpenMatch~\cite{openmatch}&60.16&62.56\\\hline
KD~\cite{hinton2015distilling}  &66.56 &67.03\\
\bf {\shortname}   &\bf{68.23} &\bf{68.93}\\\hline
Teacher &69.10  &69.10\\ \hline
\end{tabular}
\label{tab:imagenetsub}}
\vspace{-4mm}
\end{table}

\section{Conclusion}
In this work, we have presented a novel {\em \modelname{}} (\shortname) method for structured representational knowledge extraction and transfer. 
The key idea is that we take pretrained teacher's classifier as a semantic critic
for inducing a cross-network logit on student's representation.
Considering seen classes as a basis of the semantic space,
we further scale \shortname{} to highly unconstrained unlabeled data
with arbitrary unseen classes involved.
This results in a crossing of knowledge distillation and open-set semi-supervised learning (SSL).
Extensive experiments on a wide variety of network architectures and vision applications validate the performance advantages of our \shortname{}
over both state-of-the-art distillation and SSL alternatives,
often by a large margin.
Crucially, we reveal hidden limitations 
of existing open-set SSL methods in tackling more unconstrained unlabeled data,
and suggest a favor of knowledge distillation over our-of-distribution data detection.

\bibliographystyle{spmpsci}      
\bibliography{egbib}   

\begin{thebibliography}{10}
\providecommand{\url}[1]{{#1}}
\providecommand{\urlprefix}{URL }
\expandafter\ifx\csname urlstyle\endcsname\relax
  \providecommand{\doi}[1]{DOI~\discretionary{}{}{}#1}\else
  \providecommand{\doi}{DOI~\discretionary{}{}{}\begingroup \urlstyle{rm}\Url}\fi

\bibitem{ahn2019variational}
Ahn, S., Hu, S.X., Damianou, A., Lawrence, N.D., Dai, Z.: Variational information distillation for knowledge transfer.
\newblock In: IEEE Conference on Computer Vision and Pattern Recognition (2019)

\bibitem{berthelot2019mixmatch}
Berthelot, D., Carlini, N., Goodfellow, I., Papernot, N., Oliver, A., Raffel, C.A.: Mixmatch: A holistic approach to semi-supervised learning.
\newblock In: Advances on Neural Information Processing Systems (2019)

\bibitem{bulat2019xnor}
Bulat, A., Tzimiropoulos, G.: {XNOR-Net++}: Improved binary neural networks.
\newblock In: British Machine Vision Conference (2019)

\bibitem{chen2018mobilefacenets}
Chen, S., Liu, Y., Gao, X., Han, Z.: Mobilefacenets: Efficient cnns for accurate real-time face verification on mobile devices.
\newblock In: Chinese Conference on Biometric Recognition (2018)

\bibitem{chen2020simple}
Chen, T., Kornblith, S., Norouzi, M., Hinton, G.: A simple framework for contrastive learning of visual representations.
\newblock In: International Conference on Machine Learning (2020)

\bibitem{chen2020semi}
Chen, Y., Zhu, X., Li, W., Gong, S.: Semi-supervised learning under class distribution mismatch.
\newblock In: AAAI Conference on Artificial Intelligence (2020)

\bibitem{Cho_2019_ICCV}
Cho, J.H., Hariharan, B.: On the efficacy of knowledge distillation.
\newblock In: IEEE International Conference on Computer Vision (2019)

\bibitem{infotheorybook2006thomas}
Cover, T.M., Thomas, J.A.: Elements of Information Theory (Wiley Series in Telecommunications and Signal Processing).
\newblock Wiley-Interscience, USA (2006)

\bibitem{arcface}
Deng, J., Guo, J., Xue, N., Zafeiriou, S.: Arcface: Additive angular margin loss for deep face recognition.
\newblock In: IEEE Conference on Computer Vision and Pattern Recognition (2019)

\bibitem{dosovitskiy2020image}
Dosovitskiy, A., Beyer, L., Kolesnikov, A., Weissenborn, D., Zhai, X., Unterthiner, T., Dehghani, M., Minderer, M., Heigold, G., Gelly, S., et~al.: An image is worth 16x16 words: Transformers for image recognition at scale.
\newblock In: International Conference on Learning Representations (2020)

\bibitem{du2018power}
Du, S.S., Lee, J.D.: On the power of over-parametrization in neural networks with quadratic activation.
\newblock In: International Conference on Machine Learning (2018)

\bibitem{guo2020safe}
Guo, L.Z., Zhang, Z.Y., Jiang, Y., Li, Y.F., Zhou, Z.H.: Safe deep semi-supervised learning for unseen-class unlabeled data.
\newblock In: International Conference on Machine Learning (2020)

\bibitem{guo2016ms}
Guo, Y., Zhang, L., Hu, Y., He, X., Gao, J.: Ms-celeb-1m: A dataset and benchmark for large-scale face recognition.
\newblock In: European Conference on Computer Vision (2016)

\bibitem{han2015deep}
Han, S., Mao, H., Dally, W.J.: Deep compression: Compressing deep neural networks with pruning, trained quantization and huffman coding.
\newblock arXiv  (2015)

\bibitem{he2020momentum}
He, K., Fan, H., Wu, Y., Xie, S., Girshick, R.: Momentum contrast for unsupervised visual representation learning.
\newblock In: IEEE Conference on Computer Vision and Pattern Recognition (2020)

\bibitem{He_2016_CVPR}
He, K., Zhang, X., Ren, S., Sun, J.: Deep residual learning for image recognition.
\newblock In: IEEE Conference on Computer Vision and Pattern Recognition (2016)

\bibitem{heo2019comprehensive}
Heo, B., Kim, J., Yun, S., Park, H., Kwak, N., Choi, J.Y.: A comprehensive overhaul of feature distillation.
\newblock In: IEEE International Conference on Computer Vision (2019)

\bibitem{heo2019knowledge}
Heo, B., Lee, M., Yun, S., Choi, J.Y.: Knowledge transfer via distillation of activation boundaries formed by hidden neurons.
\newblock In: AAAI Conference on Artificial Intelligence (2019)

\bibitem{hinton2015distilling}
Hinton, G., Vinyals, O., Dean, J.: Distilling the knowledge in a neural network.
\newblock arXiv  (2015)

\bibitem{howard2017mobilenets}
Howard, A.G., Zhu, M., Chen, B., Kalenichenko, D., Wang, W., Weyand, T., Andreetto, M., Adam, H.: {MobileNets}: Efficient convolutional neural networks for mobile vision applications.
\newblock arXiv  (2017)

\bibitem{T2T}
Huang, J., Fang, C., Chen, W., Chai, Z., Wei, X., Wei, P., Lin, L., Li, G.: Trash to treasure: Harvesting {OOD} data with cross-modal matching for open-set semi-supervised learning.
\newblock In: IEEE International Conference on Computer Vision (2021)

\bibitem{huang2017like}
Huang, Z., Wang, N.: Like what you like: Knowledge distill via neuron selectivity transfer.
\newblock arXiv  (2017)

\bibitem{huang2021universal}
Huang, Z., Xue, C., Han, B., Yang, J., Gong, C.: Universal semi-supervised learning.
\newblock In: Thirty-Fifth Conference on Neural Information Processing Systems (2021)

\bibitem{iscen2019label}
Iscen, A., Tolias, G., Avrithis, Y., Chum, O.: Label propagation for deep semi-supervised learning.
\newblock In: IEEE Conference on Computer Vision and Pattern Recognition (2019)

\bibitem{jain2019quest}
Jain, H., Gidaris, S., Komodakis, N., P{\'e}rez, P., Cord, M.: {QUEST}: Quantized embedding space for transferring knowledge.
\newblock In: European Conference on Computer Vision (2020)

\bibitem{kemelmacher2016megaface}
Kemelmacher-Shlizerman, I., Seitz, S.M., Miller, D., Brossard, E.: The megaface benchmark: 1 million faces for recognition at scale.
\newblock In: IEEE Conference on Computer Vision and Pattern Recognition (2016)

\bibitem{kim2018paraphrasing}
Kim, J., Park, S., Kwak, N.: Paraphrasing complex network: Network compression via factor transfer.
\newblock In: Advances on Neural Information Processing Systems (2018)

\bibitem{krizhevsky2009learning}
Krizhevsky, A., et~al.: Learning multiple layers of features from tiny images.
\newblock Tech Report  (2009)

\bibitem{laine2016temporal}
Laine, S., Aila, T.: Temporal ensembling for semi-supervised learning.
\newblock International Conference on Learning Representations  (2017)

\bibitem{lan2018knowledge}
Lan, X., Zhu, X., Gong, S.: Knowledge distillation by on-the-fly native ensemble.
\newblock In: Advances on Neural Information Processing Systems (2018)

\bibitem{le2015tiny}
Le, Y., Yang, X.: Tiny imagenet visual recognition challenge.
\newblock CS 231N  (2015)

\bibitem{lebedev2016fast}
Lebedev, V., Lempitsky, V.: Fast convnets using group-wise brain damage.
\newblock In: IEEE Conference on Computer Vision and Pattern Recognition (2016)

\bibitem{lee2013pseudo}
Lee, D.H.: Pseudo-label: The simple and efficient semi-supervised learning method for deep neural networks.
\newblock In: International Conference on Machine Learning Workshop (2013)

\bibitem{lee2018self}
Lee, S.H., Kim, D.H., Song, B.C.: Self-supervised knowledge distillation using singular value decomposition.
\newblock In: European Conference on Computer Vision (2018)

\bibitem{lilocal}
Li, X., Wu, J., Fang, H., Liao, Y., Wang, F., Qian, C.: Local correlation consistency for knowledge distillation.
\newblock In: European Conference on Computer Vision (2020)

\bibitem{liu2018darts}
Liu, H., Simonyan, K., Yang, Y.: Darts: Differentiable architecture search.
\newblock International Conference on Learning Representations  (2019)

\bibitem{liu2019knowledge}
Liu, Y., Cao, J., Li, B., Yuan, C., Hu, W., Li, Y., Duan, Y.: Knowledge distillation via instance relationship graph.
\newblock In: IEEE Conference on Computer Vision and Pattern Recognition (2019)

\bibitem{martinez2020training}
Martinez, B., Yang, J., Bulat, A., Tzimiropoulos, G.: Training binary neural networks with real-to-binary convolutions.
\newblock In: International Conference on Learning Representations (2020)

\bibitem{miyato2018virtual}
Miyato, T., Maeda, S.i., Koyama, M., Ishii, S.: Virtual adversarial training: a regularization method for supervised and semi-supervised learning.
\newblock IEEE Transactions of Pattern Analysis and Machine Intelligence  (2018)

\bibitem{oliver2018realistic}
Oliver, A., Odena, A., Raffel, C.A., Cubuk, E.D., Goodfellow, I.: Realistic evaluation of deep semi-supervised learning algorithms.
\newblock Advances in neural information processing systems  (2018)

\bibitem{park2019relational}
Park, W., Kim, D., Lu, Y., Cho, M.: Relational knowledge distillation.
\newblock In: IEEE Conference on Computer Vision and Pattern Recognition (2019)

\bibitem{passalis2018learning}
Passalis, N., Tefas, A.: Learning deep representations with probabilistic knowledge transfer.
\newblock In: European Conference on Computer Vision (2018)

\bibitem{passalis2020heterogeneous}
Passalis, N., Tzelepi, M., Tefas, A.: Heterogeneous knowledge distillation using information flow modeling.
\newblock In: IEEE Conference on Computer Vision and Pattern Recognition (2020)

\bibitem{paszke2017automatic}
Paszke, A., Gross, S., Chintala, S., Chanan, G., Yang, E., DeVito, Z., Lin, Z., Desmaison, A., Antiga, L., Lerer, A.: Automatic differentiation in {PyTorch}  (2017)

\bibitem{peng2019correlation}
Peng, B., Jin, X., Liu, J., Zhou, S., Wu, Y., Liu, Y., Li, D., Zhang, Z.: Correlation congruence for knowledge distillation.
\newblock In: IEEE International Conference on Computer Vision (2019)

\bibitem{rastegari2016xnor}
Rastegari, M., Ordonez, V., Redmon, J., Farhadi, A.: {XNOR-Net: ImageNet} classification using binary convolutional neural networks.
\newblock In: European Conference on Computer Vision (2016)

\bibitem{rizve2022towards}
Rizve, M.N., Kardan, N., Shah, M.: Towards realistic semi-supervised learning.
\newblock In: European Conference on Computer Vision (2022)

\bibitem{romero2015fitnets}
Romero, A., Kahou, S.E., Montréal, P., Bengio, Y., Montréal, U.D., Romero, A., Ballas, N., Kahou, S.E., Chassang, A., Gatta, C., Bengio, Y.: Fitnets: Hints for thin deep nets.
\newblock In: International Conference on Learning Representations (2015)

\bibitem{ILSVRC15}
Russakovsky, O., Deng, J., Su, H., Krause, J., Satheesh, S., Ma, S., Huang, Z., Karpathy, A., Khosla, A., Bernstein, M., Berg, A.C., Fei-Fei, L.: Imagenet large scale visual recognition challenge.
\newblock International Journal on Computer Vision  (2015)

\bibitem{openmatch}
Saito, K., Kim, D., Saenko, K.: {OpenMatch}: Open-set consistency regularization for semi-supervised learning with outliers.
\newblock In: Advances on Neural Information Processing Systems (2021)

\bibitem{sajjadi2016regularization}
Sajjadi, M., Javanmardi, M., Tasdizen, T.: Regularization with stochastic transformations and perturbations for deep semi-supervised learning.
\newblock In: Advances on Neural Information Processing Systems (2016)

\bibitem{mobilenetv2}
Sandler, M., Howard, A., Zhu, M., Zhmoginov, A., Chen, L.C.: {MobileNetV2}: Inverted residuals and linear bottlenecks.
\newblock In: IEEE Conference on Computer Vision and Pattern Recognition (2018)

\bibitem{sharma2018conceptual}
Sharma, P., Ding, N., Goodman, S., Soricut, R.: Conceptual captions: A cleaned, hypernymed, image alt-text dataset for automatic image captioning.
\newblock In: Proceedings of the 56th Annual Meeting of the Association for Computational Linguistics (2018)

\bibitem{shi2018transductive}
Shi, W., Gong, Y., Ding, C., Tao, Z.M., Zheng, N.: Transductive semi-supervised deep learning using min-max features.
\newblock In: European Conference on Computer Vision (2018)

\bibitem{sohn2020fixmatch}
Sohn, K., Berthelot, D., Carlini, N., Zhang, Z., Zhang, H., Raffel, C.A., Cubuk, E.D., Kurakin, A., Li, C.L.: Fixmatch: Simplifying semi-supervised learning with consistency and confidence.
\newblock Advances in Neural Information Processing Systems  (2020)

\bibitem{soltanolkotabi2018theoretical}
Soltanolkotabi, M., Javanmard, A., Lee, J.D.: Theoretical insights into the optimization landscape of over-parameterized shallow neural networks.
\newblock IEEE Transactions on Information Theory  (2018)

\bibitem{tarvainen2017mean}
Tarvainen, A., Valpola, H.: Mean teachers are better role models: Weight-averaged consistency targets improve semi-supervised deep learning results.
\newblock In: Advances on Neural Information Processing Systems (2017)

\bibitem{thomee2016yfcc100m}
Thomee, B., Shamma, D.A., Friedland, G., Elizalde, B., Ni, K., Poland, D., Borth, D., Li, L.J.: {YFCC100M}: The new data in multimedia research.
\newblock Communications of the ACM  (2016)

\bibitem{tian2019contrastive}
Tian, Y., Krishnan, D., Isola, P.: Contrastive representation distillation.
\newblock In: International Conference on Learning Representations (2020)

\bibitem{tung2019similarity}
Tung, F., Mori, G.: Similarity-preserving knowledge distillation.
\newblock In: IEEE International Conference on Computer Vision (2019)

\bibitem{wang2021knowledge}
Wang, L., Yoon, K.J.: Knowledge distillation and student-teacher learning for visual intelligence: A review and new outlooks.
\newblock IEEE transactions on pattern analysis and machine intelligence  (2021)

\bibitem{wu2016quantized}
Wu, J., Leng, C., Wang, Y., Hu, Q., Cheng, J.: Quantized convolutional neural networks for mobile devices.
\newblock In: IEEE Conference on Computer Vision and Pattern Recognition (2016)

\bibitem{wu2018light}
Wu, X., He, R., Sun, Z., Tan, T.: A light {CNN} for deep face representation with noisy labels.
\newblock IEEE Transactions on Information Forensics and Security  (2018)

\bibitem{SSKD}
Xu, G., Liu, Z., Li, X., Loy, C.C.: Knowledge distillation meets self-supervision.
\newblock In: European Conference on Computer Vision (2020)

\bibitem{srrl}
Yang, J., Martinez, B., Bulat, A., Tzimiropoulos, G.: Knowledge distillation via softmax regression representation learning.
\newblock In: International Conference on Learning Representations (2021)

\bibitem{yim2017gift}
Yim, J., Joo, D., Bae, J., Kim, J.: A gift from knowledge distillation: Fast optimization, network minimization and transfer learning.
\newblock In: IEEE Conference on Computer Vision and Pattern Recognition (2017)

\bibitem{yu2020multi}
Yu, Q., Ikami, D., Irie, G., Aizawa, K.: Multi-task curriculum framework for open-set semi-supervised learning.
\newblock In: European Conference on Computer Vision (2020)

\bibitem{wrn}
Zagoruyko, S., Komodakis, N.: Wide residual networks.
\newblock In: British Machine Vision Conference (2016)

\bibitem{zagoruyko2016wide}
Zagoruyko, S., Komodakis, N.: Wide residual networks.
\newblock In: British Machine Vision Conference (2016)

\bibitem{zagoruyko2016paying}
Zagoruyko, S., Komodakis, N.: Paying more attention to attention: Improving the performance of convolutional neural networks via attention transfer.
\newblock In: International Conference on Learning Representations (2017)

\bibitem{zhang2018deep}
Zhang, Y., Xiang, T., Hospedales, T.M., Lu, H.: Deep mutual learning.
\newblock In: IEEE Conference on Computer Vision and Pattern Recognition (2018)

\bibitem{places}
Zhou, B., Lapedriza, A., Khosla, A., Oliva, A., Torralba, A.: Places: A 10 million image database for scene recognition.
\newblock IEEE Transactions of Pattern Analysis and Machine Intelligence  (2017)

\bibitem{zhou2018rocket}
Zhou, G., Fan, Y., Cui, R., Bian, W., Zhu, X., Gai, K.: Rocket launching: A universal and efficient framework for training well-performing light net.
\newblock In: AAAI Conference on Artificial Intelligence (2018)

\bibitem{zoph2016neural}
Zoph, B., Le, Q.V.: Neural architecture search with reinforcement learning.
\newblock In: International Conference on Learning Representations (2017)

\end{thebibliography}

\end{document}